\documentclass[preprint,review,12pt]{elsarticle}

\usepackage{amssymb}
\usepackage{amsfonts,graphicx,bm,amssymb,bbding}
\usepackage[fleqn]{amsmath}
\usepackage{algorithmic}
\usepackage{booktabs}
\usepackage{multirow}
\usepackage{url}
\usepackage{algorithm}
\usepackage{array}
\usepackage[caption=false,font=normalsize,labelfont=sf,textfont=sf]{subfig}
\usepackage{textcomp}
\usepackage{stfloats}
\usepackage[colorlinks,linkcolor=blue]{hyperref}
\usepackage{url}
\usepackage{verbatim}
\usepackage{graphicx}

\usepackage{soul}

\journal{Knowledge-Based Systems}

\begin{document}

\begin{frontmatter}


%%\cortext[cor1]{}

\title{LGN-Net: Local-Global Normality Network for Video Anomaly Detection}


\author[1]{Mengyang Zhao}
\ead{myzhao20@fudan.edu.cn}
\author[1]{Xinhua Zeng\corref{cor1}}
\cortext[cor1]{Corresponding author}
\ead{zengxh@fudan.edu.cn}
\author[1]{Yang Liu}
\ead{yang\_liu20@fudan.edu.cn}
\author[1]{Jing Liu}
\ead{jingliu19@fudan.edu.cn}
\author[4]{Di Li}
\ead{lidi@stu.haust.edu.cn}
\author[3]{Xing Hu}
\ead{huxing@usst.edu.cn}
\author[2]{Chengxin Pang}
\ead{chengxin.pang@shiep.edu.cn}
%%\affiliation[1]{organization={Academy for Engineering and Technology, Fudan University},%Department and Organization
%%            addressline={No.220 Handan Road}, 
%%             city={Shanghai},
%%             postcode={200433}, 
            %%state={Shanghai},
%%             country={China}}

\address[1]{Academy for Engineering and Technology, Fudan University,%Department and Organization
            No.220 Handan Road, 
           Shanghai,
            200433, 
            %%state={Shanghai},
            China}
\address[2]{School of Electronics and Information Engineering, Shanghai University of Electric Power,%Department and Organization
           No.1851 Huchenghuan Road, 
            Shanghai,
            201306, 
            %%state={Shanghai},
           China}

\address[3]{School of Optical Electrical and Computer Engineering, University of Shanghai for Science and Technology,%Department and Organization
            No.516 Jungong Road, 
            Shanghai,
           200093, 
            %%state={Shanghai},
            China}
\address[4]{College of Information Engineering, Henan University of Science and Technology,%Department and Organization
            No.263, Kaiyuan Road, 
            Luoyang,
            471000, 
           Henan,
           China}

\begin{abstract}
Video anomaly detection (VAD) has been intensively studied for years because of its potential applications in intelligent video systems. Existing unsupervised VAD methods tend to learn normality from training sets consisting of only normal videos and regard instances deviating from such normality as anomalies. However, they often consider only local or global normality in the temporal dimension. Some of them focus on learning local spatiotemporal representations from consecutive frames to enhance the representation for normal events. But powerful representation allows these methods to represent some anomalies and causes miss detection.
In contrast, the other methods are devoted to memorizing prototypical normal patterns of whole training videos to weaken the generalization for anomalies, which also restricts them from representing diverse normal patterns and causes false alarm. To this end, we propose a two-branch model, Local-Global Normality Network (LGN-Net), to simultaneously learn local and global normality. Specifically, one branch learns the evolution regularities of appearance and motion from consecutive frames as local normality utilizing a spatiotemporal prediction network, while the other branch memorizes prototype features of the whole videos as global normality by a memory module. LGN-Net achieves a balance of representing normal and abnormal instances by fusing local and global normality. In addition, the fused normality enables LGN-Net to generalize to various scenes more than exploiting single normality. Experiments demonstrate the effectiveness and superior performance of our method. The code is available online: \href{https://github.com/Myzhao1999/LGN-Net}{https://github.com/Myzhao1999/LGN-Net}.
\end{abstract}



\begin{highlights}

\item 
We introduce a novel method to simultaneously learn the local and global normality of training videos to detect anomalies in videos, which solves the inherent problem that existing methods have difficulty balancing their representation capacity for normal and abnormal patterns.
\item 
We propose a two-branch model named LGN-Net, where one branch learns the normal evolution of appearance and motion in video clips as the local normality, and the other branch memorizes normal prototype features from the whole training set as the global normality.
\item We utilize a unified cell to capture spatiotemporal representation to learn the local normality of appearance and motion rather than relying on optical flow as in previous work.
\item We fuse the local and global normality, which enables LGN-Net with excellent representation for normal patterns while limiting LGN-Net to represent anomalies.
\item The experiments on benchmark datasets indicate that our method is more adaptable to different datasets compared with state-of-the-art methods.
\end{highlights}

\begin{keyword}
Video anomaly detection \sep spatiotemporal representation \sep memory network \sep normality learning


%% or \MSC[2008] code \sep code (2000 is the defashort term

\end{keyword}

\end{frontmatter}


\section{Introduction}
Video anomaly detection (VAD) aims to identify frames where abnormal events occur. It has been intensively studied because of its potential to be used in intelligent surveillance video systems  \cite{kbs-echo,kbs-2022,muti-task,pami2021,insai}. However, it is still a challenging task due to the unbounded and rare nature of anomalies  \cite{liu-pami}. In other words, collecting all types of abnormal events is expensive or impossible. In a real application scenario, models will be challenged by various abnormal events that have never occurred before. Therefore, it is impractical to address the VAD task using a binary classification approach. 

To this end, the intuitive idea is to consider anomalies as unexpected instances or instances deviating from normal patterns. For example, a vehicle appears on a sidewalk, and a robbery occurs on campus. These events are distinct from the normal behaviors or are unexpected in that scene. Based on this perspective, many unsupervised methods have been proposed with remarkable achievements in recent years \cite{Anopcn,chen2021nm,tmm-recon,liu2022collaborative,liu2022learning,liu2022appearance}. These methods typically perform a proxy task to reconstruct input frames or predict future frames to learn the normality from training sets consisting of only normal videos. 
In the testing phase, they assume that abnormal instances will receive high reconstruction or prediction errors and be distinguished due to deviating from the learned normality.  Prediction-based methods have achieved superior performance generally and have received more focus in recent years.

\begin{figure*}
  \centering
  \includegraphics[width=.75\textwidth]{ 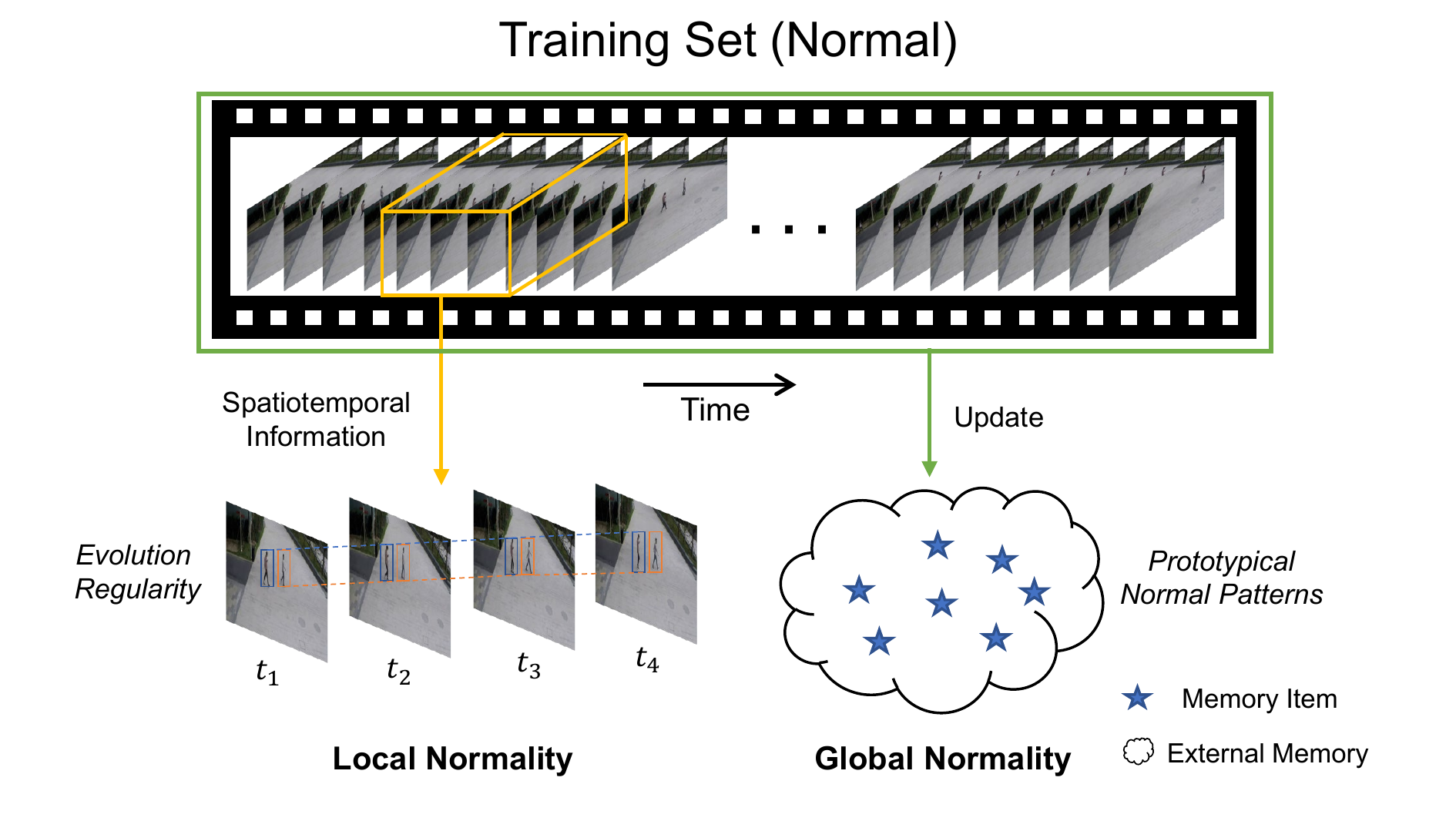}
  \caption{The local normality focuses on learning the short-term evolution regularity of normal events using spatiotemporal information from consecutive frames.  The global normality focuses on memorizing the prototypical normal patterns by updating the memory items recorded in external memory with the entire training set.
  } 
  \label{fig:lg}
\end{figure*}

In this paper, we propose to consider the short-term evolution regularity of normal events as the local normality while considering the prototypical normal pattern of the whole training set as the global normality as shown in Fig.~\ref{fig:lg}.
However, we find that existing prediction-based VAD methods typically consider only local or global normality, which causes them to have difficulty balancing their representation for normal and abnormal instances. In addition, the single normality is also difficult to generalize to various scenes because of the diversity of normal and abnormal patterns. Many existing prediction-based methods \cite{liu2018future,icassp,pr-consistency,tcsv2021} over-focus on local normality. They tend to capture spatiotemporal representations from the current input frame sequence utilizing delicately designed prediction models. According to the spatiotemporal representations, they can learn the evolution regularities of consecutive frames and consider the regularities as local normality to predict future frames. For example, Frame-Pred \cite{liu2018future} uses a fine-tuned U-Net \cite{unet} to capture the spatial representations of four consecutive frames while considering optical flow as temporal information to generate the fifth frame. In addition, adversarial training is added to Frame-Pred to generate more realistic predicted frames. Therefore, the methods over-focusing on local normality often have excellent prediction ability and can represent more diverse and complex normal patterns. However, the powerful prediction ability also allows them to predict some anomalies well and cause miss detection \cite{park2020learning}. In some simper scenes, abnormal events may be easily predicted due to their simple evolution regularities, leading to more miss detection. 

In contrast, some prediction-based methods \cite{park2020learning,ammc} over-focus on global normality. In the training phase, they attempt to record the prototype features of all normal videos by external memories to memorize prototypical normal patterns. In the testing phase, they consider the prototypical normal patterns as the learned global normality and perform the prediction task utilizing the prototype features to restrict their prediction ability for anomalies. Although these methods can limit their representation for anomalies by global normality, they almost ignore the importance of local spatiotemporal representations. It has been revealed by many excellent works \cite{wang2017predrnn,wang2021predrnn} that local spatiotemporal representations are crucial for models to understand videos. Therefore, the methods over-focusing global normality also restrict their representation for normal instances and cause false alarm. Seriously, the false alarm may be more frequent in complex or larger datasets due to the diversity and complexity of normal patterns.

To address the above issues, we proposed a two-branch model named Local-Global Normality Network (LGN-Net) to consider both local and global normality of training videos. Specifically, one branch captures local spatiotemporal representations within a unified cell to learn the evolution regularities of appearance and motion in consecutive frames, while the other branch memorizes prototype features with a memory module and updates them using all training videos to learn the prototypical normal patterns. To fuse the local and global normality, we combine the local spatiotemporal representations and the updated prototype features to generate future frames utilizing a unified decoder. Benefiting from the two-branch strategy, LGN-Net achieves a balance that allows LGN-Net to represent diverse and complex normal patterns while restricting its prediction for anomalies. In addition, the fused normality also enables LGN-Net more generalized to various scenes than exploiting single normality. 

Different from existing methods that take optical flow as motion information \cite{ammc,liu2018future,hybrid}, we capture appearance and motion evolution from video frames within a unified cell in the spatiotemporal branch inspired by PredRNN \cite{wang2017predrnn} and PredRNN-V2 \cite{wang2021predrnn}. Therefore, the spatiotemporal branch frees our model from relying on optical flow networks and reduces the computational cost. In addition, existing VAD methods \cite{liu2018future,tcsvt2020,pr-consistency} usually utilize U-Net \cite{unet} or deep autoencoder to capture spatiotemporal representation. However, these backbone networks are not adept at processing temporal sequence data. To this end, we design the spatiotemporal branch based on the recurrent neural network, which enables our model to learn temporal variation. Unlike MemAE \cite{gong2019memorizing} and LMC-Net \cite{lmc-net}, we utilize a memory module with new update and query schemes in the prototype branch inspired by MNAD \cite{park2020learning} to avoid excessive restrictions on representing normal patterns.

The contributions of this paper are summarized as follows:
\begin{itemize}
\item We introduce a novel VAD method to simultaneously consider the local and global normality of training videos in the temporal dimension, which solves the inherent problem that existing unsupervised VAD methods have difficulty balancing their representation capacity for normal and abnormal patterns.

\item We propose a two-branch model named LGN-Net, where one branch learns the local normality of appearance and motion using a unified cell rather than relying on optical flow as in previous work, and the other branch memorizes global prototype features from the whole training video with a memory module. The two-branch strategy enables LGN-Net with excellent representation for normal patterns while limiting LGN-Net to represent anomalies.

 \item The experiment results indicate that our method is more adaptable to various scenes with different complexity. In addition, we provide a comprehensive experimental analysis to validate the effectiveness of our method, including visualization analysis and ablation studies.
\end{itemize}
\label{}
\section{RELATED WORK}

In unsupervised VAD methods, the training sets are generally devised to contain only normal videos. In the training phase, unsupervised VAD methods tend to learn the normality from training sets by performing a proxy task. In the testing phase, any instance that significantly deviates from the learned normality will be treated as abnormal \cite{DBLP:conf/icmcs/LaiLH20}. According to the proxy task performed, the unsupervised VAD methods can be classified as reconstruction-based or prediction-based.
\subsection{Reconstruction-based Methods}

In reconstruction-based methods, models are usually designed based on autoencoder \cite{hasan2016learning,gong2019memorizing,park2020learning}. They typically utilize current input video frames as ground truth and aim to achieve a trained autoencoder that minimizes the reconstruction error. Due to only learning how to reconstruct normal frames in the training phase, they assume that the reconstruction errors of abnormal frames will be higher than that of normal frames in the testing phase. In recent years, many reconstruction-based VAD methods have been proposed.
Hasan \textit{et al.} \cite{hasan2016learning} proposed to learn normal regularity in video frames using an autoencoder based on the deep convolution neural network.
Luo \textit{et al.} \cite{luo2017revisit} proposed a temporally-coherent sparse coding to force neighboring frames reconstructed by similar reconstruction coefficients. Ravanbakhsh \textit{et al.} \cite{abnormalgan} introduced a VAD model based on Generative Adversarial Network (GAN) \cite{goodfellow2014generative}. They use normal frames to reconstruct corresponding optical flow images and use optical flow images to reconstruct corresponding normal frames. Compared with traditional VAD methods, reconstruction-based methods have made significant progress. However, due to the powerful representation of convolution neural networks, they can sometimes reconstruct the abnormal frames with low errors. To this end, Gong \textit{et al.} \cite{gong2019memorizing} devised a memory module to memorize the normal prototypical elements inspired by memory network \cite{memory-network}, constraining the model's representation for abnormal frames. Reconstruction-based methods have shown some good performance for the VAD task,  but they almost ignore the temporal representation.

\subsection{Prediction-based Methods}
In prediction-based methods \cite{liu2018future, tcsvt2020}, the models usually take several consecutive video frames as input to generate the next frame. Prediction-based methods are more adept at exploiting spatiotemporal information among frames than reconstruction-based methods. Therefore, they are usually superior to reconstruction-based methods and have received more attention recently. 
Liu \textit{et al.} \cite{liu2018future} use four consecutive frames as input to predict the fifth frame. They utilized the fine-tuned U-Net \cite{unet} as a generator to generate future frames and devise a discriminator to enhance the model's prediction ability. In addition, they take optical flow as motion (temporal) information to help the model generate more realistic future frames. Lee \textit{et al.}  \cite{icassp} proposed a VAD model named STAN, which consists of a ConvLSTM-based \cite{xingjian2015convolutional} generator and a spatiotemporal discriminator. Park \textit{et al.} \cite{park2020learning} devised a new memory module for the VAD model to record normal prototype features, which can represent more diverse patterns than the memory module in MemAE \cite{gong2019memorizing}. They use normal prototype features to represent predicted frames, which limit the prediction ability for abnormal events. Recently, many two-stream models have been proposed \cite{chang2022video,ammc,pr-consistency}, which often utilize two branches to learn spatial and temporal normal patterns separately. Chang \textit{et al.} \cite{chang2022video} devised a two-stream model consisting of a spatial autoencoder and a temporal autoencoder to dissociate the spatiotemporal representations. The model performs the reconstruction task using the spatial autoencoder to learn normal appearance patterns while performing the prediction task using the temporal autoencoder based on U-Net to learn normal motion patterns. Cai \textit{et al.} \cite{ammc} attempted to explore spatiotemporal consistency and proposed a two-stream model named AMMC-Net. They designed two autoencoders with memory modules, one with frames as input for learning normal appearance patterns and the other with the optical flow as input for learning normal motion patterns. Hao \textit{et al.} \cite{pr-consistency} introduce a model named STCEN, composed of a 3D CNN-based encoder and a 2D CNN-based decoder, to extract motion and appearance fusion features. In addition, STCEN uses a 3D CNN-based discriminator to make predicted frames more realistic.

Although the above methods have made efforts to learn spatiotemporal representations and achieve better performance, they are almost incapable of taking into account both local and global normality of training videos. Many of them \cite{liu2018future,icassp,pr-consistency,chang2022video} focus on learning local normality and enhancing prediction ability by exploiting appearance and motion information or devising a discriminator for the generative model. The powerful prediction ability enables them to represent some abnormal instances well. Besides, many prediction-based methods \cite{park2020learning,ammc} tend to record prototypical normal patterns while ignoring local spatiotemporal representation. 
\section{The Proposed Method}

Anomalies in videos can be defined as unexpected behavior or behavior deviating from normal patterns \cite{ramachandra2020survey}. Based on this definition, we introduce an unsupervised VAD method aiming to learn both local and global normality from training sets consisting of only normal videos by performing a prediction task. The instances deviating from the learned normality will achieve high prediction errors and thus are regarded as anomalies. Specifically, we propose a two-branch model, where one branch aims to learn the evolution regularities of appearance and motion from several consecutive frames and considers the regularities in this short term as local normality while the other branch updates the prototype features recorded in a memory pool using all training videos and consider the prototype features as global normality.

\begin{figure*}
  \centering
  \includegraphics[width=1.0\textwidth]{ 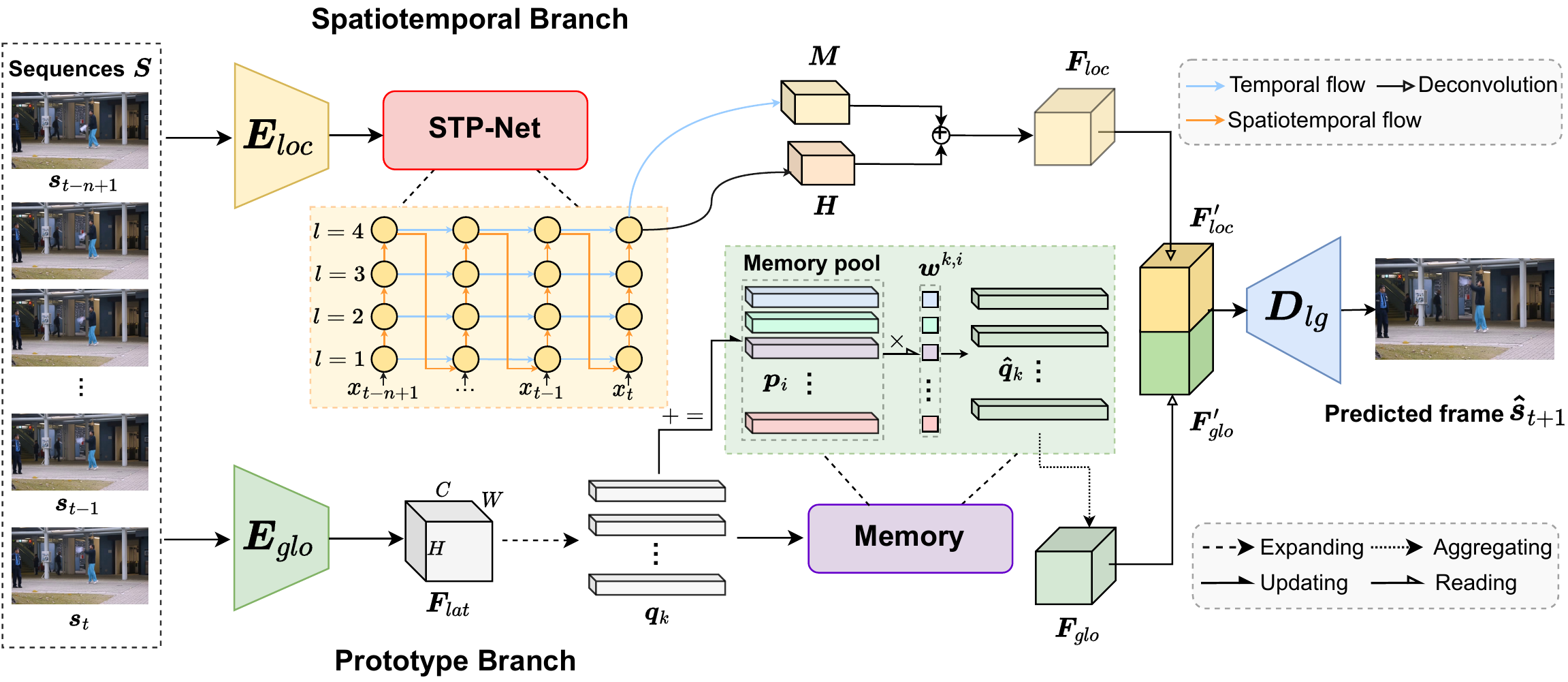}
  \caption{The overall structure of LGN-Net. STP-Net consists of four ST-LSTM \cite{wang2017predrnn} units to model the short-term evolution of appearance and motion without optical flow. The memory module aims to update the prototype features $\bm p _i$ recorded in the memory pool using all normal videos to learn global normality. $ w^{k,i}$ is a matching probability  calculated based on the distance between $\bm q_k$ and $\bm p_i$. 
  } 
  \label{LGN-Net}
\end{figure*}

\subsection{Overall Framework}
As shown in Fig.~\ref{LGN-Net}, the proposed LGN-Net consists of a spatiotemporal branch and a prototype branch. We denote the video frame sequence as $\bm{S}=\left\{\bm{ s}_{t-n+1}, \bm{ s}_{t-n+2},  \ldots, \bm{ s}_{t}\right\}$, which contains $n$ consecutive video frames. In the spatiotemporal branch, we first feed the sequence $\bm S$ to the encoder $E_{loc}$ and denote these frames in $\bm S$ after encoding as $\bm \left\{\bm{ x}_{t-n+1}, \bm{ x}_{t-n+2},  \ldots, \bm{ x}_{t}\right\}$. Then, every $\bm{x}$ is input to the spatiotemporal prediction network (STP-Net) in time step order. The STP-Net is a convolutional recurrent neural network aiming to capture the spatial and temporal representations within a unified cell. We denote the final spatiotemporal memory and the final hidden state of STP-Net as $\bm{M}$ and $\bm{H}$ respectively, which contain the spatiotemporal representations from time step $t-n+1$ to $t$ in $\bm{S}$. Therefore, we concatenate $\bm{M}$ and $\bm{H}$ as the local spatiotemporal feature $\bm {F}_{loc}$ finally. In the prototype branch, we first feed the frame sequence $\bm S$ to the encoder $E_{glo}$ to generate latent space feature $\bm{F}_{lat}$. Then, $\bm{F}_{lat}$ is expanded to a number of query features, which are used to update the prototype features recorded in the memory pool. The query features are represented approximately by combining updated prototype features according to a matching probability. After the training phase, the prototype features will be updated by all sequences $\bm S$ from all training videos to memorize the prototypical normal patterns. Then, we obtain the global feature $\bm{F}_{glo}$ by aggregating the reconstructed query features to restrict LGN-Net's representation for anomalies. Finally, the local spatiotemporal feature $\bm {F}_{loc}$ and the global feature $\bm{F}_{glo}$ are concatenated after deconvolution and fed to the decoder $D_{lg}$ to generate the next frame $\bm\hat{ s}_{t+1}$.

\subsection{Spatiotemporal Branch}
We design the spatiotemporal branch for learning short-term evolution regularities of appearance and motion without optical flow and consider the regularities as local normality.
As shown in Fig.~\ref{LGN-Net}, the spatiotemporal branch mainly consists of two components: the encoder $E_{loc}$ and the spatiotemporal prediction network (STP-Net). $E_{loc}$ is designed based on 2d convolutional neural networks, which aims to obtain the latent space features $\bm {X} = \left\{\bm{ x}_{t-n+1}, \bm{ x}_{t-n+2},  \ldots, \bm{ x}_{t}\right\}$ from sequence $\bm{S}$. The STP-Net is a convolutional recurrent neural network employed to learn the local evolution regularities of appearance and motion.

\begin{figure}[htb]
  \centering
  \includegraphics[width=.7\textwidth]{ 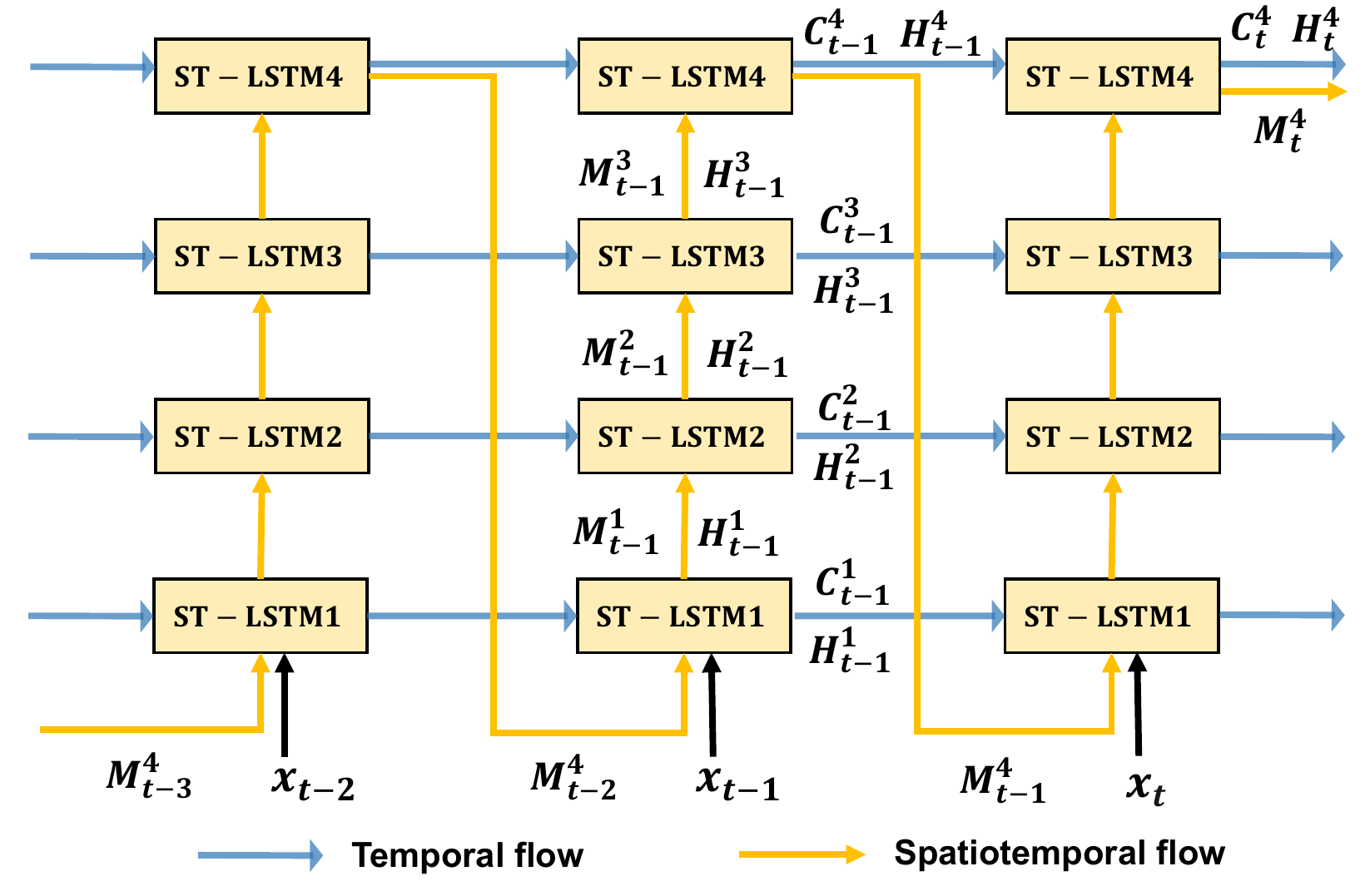}
  \caption{Details of the STP-Net. The Spatiotemporal memory is updated in the zigzag direction, enabling the bottom unit (ST-LSTM$_1$) to capture the spatial information output by the top unit (ST-LSTM$_4$) at the previous time step.}
  \label{fig:STP-Net}
\end{figure}

Compared with deep autoencoder and U-Net \cite{unet} used in previous work \cite{liu2018future,tcsvt2020,pr-consistency}, STP-Net is more adept at learning temporal variations because of its recurrent structure. Furthermore, we utilize ST-LSTM \cite{wang2017predrnn} as the unit of STP-Net because it captures spatial and temporal representations within a unified cell and achieves superior performance over ConvLSTM \cite{xingjian2015convolutional} in appearance and motion prediction.
Similar to ConvLSTM \cite{xingjian2015convolutional}, the temporal cell  $\bm{C}_{t}^{l}$ in STP-Net is updated horizontally from the previous time step to the current time step within all layers as shown in Fig.~\ref{fig:STP-Net}. In addition to $\bm{C}_{t}^{l}$, ST-LSTM devised a spatiotemporal memory cell $\bm{M}_{t}^{l}$ to model both spatial and temporal features. For the bottom of STP-Net, spatiotemporal memory conveyed from the top layer at the previous time step to the bottom layer at the current time step, denoted as $\bm{M}_{t}^{0}=\bm{M}_{t-1}^{4}$. For the other layers, spatiotemporal memory is transmitted from $l-1$ layer to the current layer vertically at the same time step. Therefore, the zigzag direction update strategy enables STP-Net to overcome the layer-independent memory mechanism in ConvLSTM \cite{xingjian2015convolutional}  and memorize the evolution regularities of appearance and motion in a unified memory cell. The key equations of STP-Net are shown as follows: 

 \begin{equation}
 \begin{aligned}
   {i}_{t} & =\sigma({W}^x_{i} \ast \bm{x}_{t}+{W}^h_{i} \ast \bm{H}_{t-1}^{l})\\
   {f}_{t} & =\sigma({W}^x_{f} \ast \bm{x}_{t}+{W}^h_{f} \ast \bm{H}_{t-1}^{l})\\
   {i}^\prime_{t} & =\sigma(W^x_{im} \ast \bm{x}_{t}+ W^m_{i} \ast \bm{M}_{t}^{l-1})\\
   f^\prime_{t} & =\sigma(W^x_{fm} \ast \bm{x}_{t}+W^m_{f} \ast \bm{M}_{t}^{l-1})\\
   \bm{C}_{t}^{l} & =f_{t} \odot \bm{C}_{t-1}^{l}+i_{t} \odot \tanh ({W}^x_{c} \ast \bm{x}_{t} +{W}^h_{c} \ast\bm{ H}_{t-1}^{l})\\
 \bm{M}_{t}^{l} & = f_{t}^{\prime} \odot \bm{M}_{t}^{l-1}+i_{t}^{\prime} \odot \tanh ({W}^x_m \ast\bm{ x}_{t}  +W^m_m \ast \bm{M}_{t}^{l-1})\\
  o_{t} & =\sigma(W^x_{o} \ast \bm{x}_{t}+W^h_{o} \ast \bm{H}_{t-1}^{l}+W^c_{o} \ast \bm{C}_{t}^{l} +W^m_{o} \ast \bm{M}_{t}^{l})\\
  \bm{H}_{t}^{l} & = o_{t} \odot \tanh \left(W_{1 \times 1} \ast\left[\bm{C}_{t}^{l}, \bm{M}_{t}^{l}\right]\right)
  \label{Eq:eq1}
   \end{aligned}
\end{equation}%

where $\ast$ and $\mathcal\odot$ denote the convolution operator and the Hadamard product, respectively. $\mathcal\sigma$ is the sigmoid activation function. ${i}_{t}$ and ${f}_{t}$ are the input gate and forget gate of the temporal cell, respectively, while ${i}^\prime_{t}$ and ${f}^\prime_{t}$ are the input gate and forget gate of the spatiotemporal cell, respectively.

As shown in Fig.~\ref{fig:STP-Net}, in order to learn the evolution regularities of appearance and motion in $\bm{S}$, we feed every $\bm{x}$ individually into STP-Net in time step order.
According to Eq.~\ref{Eq:eq1}, we can learn that the final spatiotemporal memory $\bm{M}^4_t$  contains the spatiotemporal information from the past time step to the current time step and the final hidden state $\bm{H}_{t}^{4}$ consider both standard temporal memory and spatiotemporal memory. Therefore, we concatenate $\bm{H}_{t}^{4}$ and $\bm{M}^4_t$ together as the local spatiotemporal representation $\bm {F}_{loc}$ of the current input sequence. Note that we denote $\bm{H}_{t}^{4}$ and $\bm{M}^4_t$ as $\bm{H}$ and $\bm{M}$ respectively in Fig.~\ref{LGN-Net}. The local spatiotemporal representations made it unnecessary to use optical flow as motion information, which only tends to represent variations between adjacent frames \cite{nguyen_anomaly_2020} and results in higher computational cost. LGN-Net explores complete evolution regularities of appearance and motion from input frame sequence by the spatiotemporal branch.

\subsection{Prototype Branch}
We design the prototype branch to memorize the prototype features with a memory module from whole training videos inspired by MNAD \cite{park2020learning} and consider the prototype features as global normality. 
As shown in Fig.~\ref{LGN-Net}, the prototype branch is designed based on a memory-augmented deep autoencoder. The branch has three components: the encoder $E_{glo}$, the memory module, and the decoder $D_{lg}$. The encoder $E_{glo}$ is designed based on 2d convolutional neural networks, which takes the sequence $\bm{ S}$  as input and outputs the latent space features $\bm {F}_{lat}\in \mathbb{R}^{H \times W \times C }$. The memory module records prototype features in a memory pool and updates them utilizing $\bm {F}_{lat}$ from all training videos to memorize prototypical normal patterns. Therefore, we can consider the updated prototype features as the learned global normality in the testing phase. We represent $\bm {F}_{lat}$ approximately using the updated prototype features to limit LGN-Net's representation for anomalies and denote it as $\bm{F}_{glo}$.  $\bm {F}_{glo}$ and $\bm {F}_{loc}$ have the same dimension after deconvolution, denoted as $\bm {F}^\prime_{glo}$ and $\bm {F}^\prime_{loc}$ respectively. Finally, we concatenate $\bm {F}^\prime_{glo}$ and the $\bm {F}^\prime_{loc}$ to fuse local and global normality and then feed them into the decoder $D_{lg}$ to generate a future frame $\hat{\bm{s}}_{t+1}$.

In the memory module, the latent space features $\bm {F}_{lat}$ are expanded into a number of query features, denoted as $ \bm {q}_k \in \mathbb{R}^{C}  (k=1, 2, ..., K)$, where $K=H \times W $. The memory module records a number of prototype features with the same dimension as feature  $ \bm {q}_k$, denoted as $ \bm {p}_i\in \mathbb{R}^{C} (i=1, 2, ..., I)$. The query scheme of the memory module determines how to represent query features $\bm {q}$ using the recorded prototype features $\bm {p}$ approximately, while the update scheme describes how to update $\bm {p}$ using the current $\bm {q}$.

In the query scheme, we compute the similarity between $\bm {q}$ and $\bm {p}$ using the cosine similarity and then apply a softmax function vertically to obtain the matching probability $w^{k,i}$: 
\begin{equation}
w^{k, i}=\frac{\exp \left(\left(\bm {p}_i\right)^T \bm  {q}_k\right)}{\sum_{i^{\prime}=1}^I \exp \left(\left(\bm {p}_{i^{\prime}}\right)^T \bm  {q}_k\right)}
\label{eq2}
.\end{equation}
To represent $ \bm {q}_k$ approximately, we obtain $\hat{\bm{q}}_k$ using a linear combination of all the prototype features $\bm {p}$ according to $w^{k,i}$ as shown in Eq.~\ref{eq3}. Then, all $\hat{\bm{q}}_k$ corresponding to $ \bm {q}_k$ are aggregated and denoted as the global feature $\bm{F}_{glo}$ to represent $\bm{F}_{lat}$ approximately.
\begin{equation}
\hat{\bm q}_k=\sum_{i^{\prime}=1}^I w^{k, i^{\prime}} \bm {p}_{i^{\prime}}
\label{eq3}
.\end{equation}

In the update scheme, all query features claiming to be nearest to the prototype feature according to the matching probabilities in Eq.~\ref{eq2} are selected to update it. We use $U^i$ to denote the set of query features for updating $ \bm {p}_i$. A weighted average of the query features is used to update $ \bm {p}_i$ to memorize the prototype features:
\begin{equation}
\hat{\bm p}_i=\left\|\bm {p}_i+\sum_{k \in U^i} v^{ k, i}_u \bm  {q}_k\right\|_2
.\end{equation}
Similar to Eq.~\ref{eq2}, we apply the softmax function in the horizontal direction to obtain $v^{k,i}$:
\begin{equation}
v^{k, i}=\frac{\exp \left(\left(\bm {p}_i\right)^T \bm  {q}_k\right)}{\sum_{k^{\prime}=1}^K \exp \left(\left(\bm {p}_i\right)^T \bm {q}_{k^{\prime}}\right)}
.\end{equation}
Then, $v^{k,i}$ is renormalized to consider the query features in $U^i$:
\begin{equation}
v^{ k, i}_u=\frac{v^{k, i}}{\max _{k^{\prime} \in U^i} v^{k^{\prime}, i}}
.\end{equation}

Additionally, in the testing phase, the prototype features may also be updated by some frames in testing videos, which allows our model to make adequate use of testing videos. We use a weighted regular score to prevent the prototype features from updating by abnormal instances. $R_t$ is used to denotes the regular score of frame $\bm{s}_t$:
\begin{equation}
{R}_t=\sum_{i, j} W_{i j}\left(\hat{\bm{s}}_t, \bm{s}_t\right)\left\|\hat{\bm{{s}}}_t^{i, j}-\bm{{s}}_t^{i,j}\right\|_2,
\end{equation} where the weight function $W_{i j}$ is represented as follows:
\begin{equation}
W_{i j}\left(\hat{\bm{{s}}}_t, \bm{{s}}_t\right)=\frac{1-\exp \left(-\left\|\hat{\bm{{s}}}_t^{i, j}-\bm{{s}}_t^{i, j}\right\|_2\right)}{\sum_{i, j} 1-\exp \left(-\left\|\hat{\bm{{s}}}_t^{i, j}-\bm{{s}}_t^{i, j}\right\|_2\right)},
\end{equation} and $i,j$ are the spatial indices of pixels. We set a threshold $\gamma$ and consider the frames with regular score $R_t$  higher than $\gamma$ as possible abnormal frames and do not use these frames to update the prototype features.

\subsection{Training loss}
To train our model, we introduce three loss functions: intensity loss, compactness loss, and separateness loss. 

The intensity loss $\mathcal{L}_{int}$ is calculated based on L2-norm to ensure the similarity of the pixels between a predicted frame $\hat{\bm{s}}$ and its corresponding ground truth $\bm{s}$:
\begin{equation}
\mathcal{L}_{int}(\hat{\bm{s}}, \bm{s})=\left\|\hat{\bm{s}}-\bm{s}\right\|_2
.\end{equation}

The compactness loss $\mathcal L_{com}$ aims to reduce the distance between the query feature $ \bm {q}_k$ and its closest prototype feature $\bm {p}_c$, which enables the prototype features have superior representation. The feature compactness loss is shown as follows:

\begin{equation}
\mathcal L_{com}=\sum_k^K\left\|\bm  {q}_k-\bm {p}_c\right\|_2
,\end{equation} where $c$ is the index of the prototype feature closest to $ \bm {q}_k$.

The prototype features recorded in the memory pool should be separated from each other adequately to represent diverse patterns. Therefore, we utilize the separateness loss $\mathcal L_{sep}$ in this work, defined with a margin of $\alpha$ as: 
\begin{equation}
\mathcal L_{sep}=\sum_k^K\left[\left\|\bm  {q}_k-\bm {p}_c\right\|_2-\left\|\bm  {q}_k-\bm {p}_s\right\|_2+\alpha\right]_{+}
,\end{equation}  where $s$ is the index of prototype feature second closest to $ \bm {q}_k$, and $\alpha$ is a constant. 

The final training loss $\mathcal L_{lgn}$ of our proposed model considers all the above loss functions, which can be represented as follow: 
\begin{equation}
\mathcal L_{lgn}=\mathcal L_{int}+\lambda_c \mathcal L_{com }+\lambda_s \mathcal L_{sep }
,\end{equation} where $\lambda_c$ and $\lambda_s$ are used to balance each loss in the final objective function.
\subsection{Normality Score}
In the testing phase, we quantify the extent of normality using the normality score for each frame. The normality score is usually calculated by the difference between the predicted frame and the corresponding ground truth \cite{liu2018future,icassp}. In this work, we calculate the normality score using the peak signal to noise ratio (PSNR) and the difference between query and prototype features.

PSNR has proven to be an effective measure in past work \cite{PSNR}. The PSNR between a predicted frame $\hat{\bm{s}}_t$ and its corresponding ground truth $\bm{s}_t$ can be shown as follows:
\begin{equation}
P\left(\hat{\bm{s}}_t, \bm{s}_t\right)=10 \log _{10} \frac{\max \left(\hat{\bm{s}}_t\right)}{\left\|\hat{\bm{s}}_t-\bm{s}_t\right\|_2^2 / N}
,\end{equation}
where $N$ is the number of pixels in $\bm{s}_t$.

We compute the L2 distance between each query feature $\bm {q}_k$ and its
closest prototype feature $\bm {p}_c$  as the difference, which reflects the deviation extent of the input instance from the global normality:
\begin{equation}
D\left(\bm{q}, \bm{p}\right)=\frac{1}{K} \sum_k^K\left\|\bm {q}_k-\bm{\bm {p}_c}\right\|_2
.\end{equation}

Then, we normalize PSNR and distance $D$ to [0, 1]:
\begin{equation}
{S}_t=\lambda\left(1-g\left(P\left(\hat{\bm{s}}_t, \bm{s}_t\right)\right)\right)+(1-\lambda) g\left(D\left(\bm{q}, \bm{p}\right)\right)
\label{eqd},\end{equation} where $\lambda$ is used to balance PSNR and distance $D$.  $g(\cdot)$ is a normalization function,which can be shown as follows:
\begin{equation}
g(P\left(\hat{\bm{s}}_t, \bm{s}_t\right))=1-\frac{{P}\left(\bm{s}_t, \hat{\bm s}_t\right)-\min _t {P}\left(\bm{s}_t, \hat{\bm s}_t\right)}{\max _t{P}\left(\bm{s}_t, \hat{\bm s}_t\right)-\min _t {P}\left(\bm{s}_t, \hat{\bm s}_t\right)}
.\end{equation}

Finally, we obtain the normality score $N_t$, and a higher $N_t$ means a lower prediction error:
\begin{equation}
N_t=1-{S}_t
.\end{equation}

\section{Experiments}

In order to validate the effectiveness of our method, we provide extensive experimental results and analysis in this section. 
\subsection{Evaluate Metrics and Benchmark Datasets}
\noindent \textbf{Evaluation Metrics.} To quantitatively evaluate our method, following the previous work \cite{gong2019memorizing,liu2018future,park2020learning}, we calculate the receiver operating characteristic (ROC) curve and utilize the average area under the curve (AUC) as the evaluation metrics. In this work, we use frame-level AUC to evaluate the performance of VAD methods, and the higher AUC indicates superior performance.

\noindent \textbf{Benchmark Datasets.} We performed experiments on three benchmark datasets, including UCSD Ped2 \cite{mahadevan2010anomaly}, CUHK Avenue \cite{lu2013abnormal} and ShanghaiTech \cite{luo2017revisit}, which have been used in many excellent VAD work \cite{liu-pami,liu2018future,park2020learning,tmm-crowded,tmm-end}.
\begin{itemize}
   \item \textbf{UCSD Ped2} \cite{mahadevan2010anomaly} contains 16 training and 12 testing  videos. The dataset is captured on the University of California San Diego pedestrian walkways. In the dataset, the normal and abnormal patterns are relatively simple. The dataset defines walking pedestrians as normal and treats events such as bicycling and skateboarding as abnormal.
  \item \textbf{CUHK Avenue} \cite{lu2013abnormal} dataset includes 37 videos of about 30,000 frames in total, with 21 testing videos including abnormal events. In the dataset, normal patterns are also simple similar to UCSD ped2. However, abnormal patterns in this dataset are more complex than in UCSD ped2, which contains a variety of abnormal events such as fast-running, throwing, and loitering.
  \item \textbf{ShanghaiTech} \cite{luo2017revisit} dataset consists of 330 training videos and 107 testing videos. This dataset is much more complex and larger compared with  UCSD ped2 and CUHK Avenue. This dataset contains a large amount of video data and was captured in 13 different scenes, which lead to the normal patterns in it being diverse and complex. In addition, the dataset contains a variety of complex abnormal events, such as chasing, fighting, and robbing. 
\end{itemize}
\subsection{Implementation Details}
In the experiments, all frames are resized to $256 \times 256$, and the intensity of pixels is normalized to [-1, 1]. We use Adam \cite{kingma2014adam} to optimize our model with a batch size of 8. The learning rate of LGN-Net is set to $2\times 10^ {-4} $ and $\alpha$ is set to 1. $n$ is empirically set to 4, which means we use four consecutive frames to predict the fifth frame in the prediction task. The dimension of $\bm {x}_t$ in the spatiotemporal branch is set to $64 \times 64 \times 128$. $H$, $W $, and $C $ are set to 32, 32, and 512, respectively, which means the dimension of latent space features $\bm F_{lat}$ in the prototype branch is $32 \times 32 \times 512$.   

When processing simple datasets, the representation capability of LGN-Net needs to be strictly limited to prevent predicting simple abnormal instances well. Therefore, for UCSD ped2 \cite{mahadevan2010anomaly} and CUHK Avenue \cite{lu2013abnormal}, $I$ is set to 10, which means the memory module only records ten normal prototype features. Correspondingly, for ShanghaiTech \cite{luo2017revisit}, $I$ is set to 200 to represent the complex and diverse normal patterns. For UCSD ped2, CUHK Avenue, and ShanghaiTech, $\lambda_c$ is set to 10, 10, and 1, $\lambda_s$ is set to 5, 2, and 1, $\lambda$ is set to 0.6, 0.5, and 0.8, and $\gamma$ is set to 0.009, 0.006, and 0.0135, respectively.

\subsection{Comparison with Existing Methods}
In this section, we compared our method with existing state-of-the-art methods on three benchmark VAD datasets mentioned above. The methods involved in the comparison experiment can be classified as traditional methods \cite{kim2009observe,mahadevan2010anomaly}, reconstruction-based  methods (Reconstruction) \cite{luo2017revisit,hasan2016learning,gong2019memorizing,park2020learning,abnormalgan,luo2017remembering,liu-pami}, and prediction-based methods (Prediction)  \cite{liu2018future,nguyen2019anomaly,chang2022video,ammc,park2020learning,icassp,pr-consistency,tcsv2021,tcsvt2019,stc-net,kbs-2022,tnnls-2022}. The proposed method belongs to prediction-based methods. We report the frame-level AUC (\%) of all the methods, and the comparison results are presented in Table~\ref{tab:1}.

The traditional methods in Table~\ref{tab:1} extract feature using handcraft and employ a one-class classifier to perform classification tasks. Therefore, these methods are not robust in complex scenes and large datasets. Therefore, compared with these methods, reconstruction-based and prediction-based methods achieve superior performance significantly.

\begin{table*}
\footnotesize
  \centering
  \caption{Results of quantitative frame-level AUC (\%) comparison. Bold Numbers indicate the best performance, while underlie ones indicate the second best.}
  \setlength{\tabcolsep}{2mm}{
    \begin{tabular}{rrrrrr}
\cmidrule{1-5}    \multicolumn{1}{l}{Types} & \multicolumn{1}{l}{Methods} & \multicolumn{1}{c}{UCSD Ped2} & \multicolumn{1}{c}{CUHK Avenue} & \multicolumn{1}{c}{ShanghaiTech} &  \\
\cmidrule{1-5}    \multicolumn{1}{l}{\multirow{3}[2]{*}{Traditional}} & \multicolumn{1}{l}{MPPCA \cite{kim2009observe}} & \multicolumn{1}{c}{69.3} & \multicolumn{1}{c}{N/A} & \multicolumn{1}{c}{N/A} &  \\
          & \multicolumn{1}{l}{MPCC+SFA \cite{mahadevan2010anomaly}} & \multicolumn{1}{c}{64.3} & \multicolumn{1}{c}{N/A} & \multicolumn{1}{c}{N/A} &  \\
          & \multicolumn{1}{l}{MDT \cite{mahadevan2010anomaly}} & \multicolumn{1}{c}{82.9} & \multicolumn{1}{c}{N/A} & \multicolumn{1}{c}{N/A} &  \\
\cmidrule{1-5}    \multicolumn{1}{l}{\multirow{8}[2]{*}{Reconstruction}} & \multicolumn{1}{l}{Conv-AE \cite {hasan2016learning}} & \multicolumn{1}{c}{85.0} & \multicolumn{1}{c}{80.0} & \multicolumn{1}{c}{60.9} &  \\
          & \multicolumn{1}{l}{TSC \cite{luo2017revisit} } & \multicolumn{1}{c}{91.0} & \multicolumn{1}{c}{80.6} & \multicolumn{1}{c}{67.9} &  \\
          & \multicolumn{1}{l}{StackRNN \cite{luo2017revisit} } & \multicolumn{1}{c}{92.2} & \multicolumn{1}{c}{81.7} & \multicolumn{1}{c}{68.0} &  \\
         & \multicolumn{1}{l}{AbnormalGAN \cite{abnormalgan} } & \multicolumn{1}{c}{93.5} & \multicolumn{1}{c}{N/A} & \multicolumn{1}{c}{N/A} &  \\
          & \multicolumn{1}{l}{ConvLSTM-AE \cite{luo2017remembering}} & \multicolumn{1}{c}{88.1} & \multicolumn{1}{c}{77.0} & \multicolumn{1}{c}{N/A} &  \\
           & \multicolumn{1}{l}{MemAE \cite{gong2019memorizing}} & \multicolumn{1}{c}{94.1} & \multicolumn{1}{c}{83.3} & \multicolumn{1}{c}{71.2} &  \\
          
          & \multicolumn{1}{l}{MNAD-Recon \cite{park2020learning} } & \multicolumn{1}{c}{90.2} & \multicolumn{1}{c}{82.8} & \multicolumn{1}{c}{69.8} &  \\
          & \multicolumn{1}{l}{SC-Net \cite{liu-pami} } & \multicolumn{1}{c}{92.2} & \multicolumn{1}{c}{83.5} & \multicolumn{1}{c}{69.6} &  \\
          
\cmidrule{1-5}    \multicolumn{1}{l}{\multirow{13}[2]{*}{Prediction}} 
          & \multicolumn{1}{l}{Frame-Pred \cite{liu2018future}} & \multicolumn{1}{c}{95.4} & \multicolumn{1}{c}{84.9} & \multicolumn{1}{c}{72.8} &  \\
           & \multicolumn{1}{l}{STAN \cite{icassp}} & \multicolumn{1}{c}{96.5} & \multicolumn{1}{c}{87.2} & \multicolumn{1}{c}{N/A} &  \\
          & \multicolumn{1}{l}{AMC \cite{nguyen2019anomaly}} & \multicolumn{1}{c}{96.2} & \multicolumn{1}{c}{86.9} & \multicolumn{1}{c}{N/A} &  \\
          & \multicolumn{1}{l}{MESDnet \cite{tcsvt2019}} & \multicolumn{1}{c}{96.0} & \multicolumn{1}{c}{86.0} & \multicolumn{1}{c}{N/A} &  \\
          & \multicolumn{1}{l}{Multispace \cite{tcsv2021}} & \multicolumn{1}{c}{95.4} & \multicolumn{1}{c}{86.8} & \multicolumn{1}{c}{73.6} &  \\
          & \multicolumn{1}{l}{MNAD \cite{park2020learning}} & \multicolumn{1}{c}{\ul{97.0}} & \multicolumn{1}{c}{\ul{88.5}} & \multicolumn{1}{c}{70.5} &  \\
          & \multicolumn{1}{l}{AMMC-Net \cite{ammc}} & \multicolumn{1}{c}{96.6} & \multicolumn{1}{c}{86.6} & \multicolumn{1}{c}{\ul {73.7}} &  \\
          & \multicolumn{1}{l}{STC-Net \cite{stc-net}} & \multicolumn{1}{c}{96.7 } & \multicolumn{1}{c}{87.8 } & \multicolumn{1}{c}{73.1} &  \\
          & \multicolumn{1}{l}{STD \cite{chang2022video}} & \multicolumn{1}{c}{96.7} & \multicolumn{1}{c}{87.1} & \multicolumn{1}{c}{\ul {73.7}} &  \\
          & \multicolumn{1}{l}{STCEN \cite{pr-consistency}} & \multicolumn{1}{c}{96.9} & \multicolumn{1}{c}{86.6} & \multicolumn{1}{c}{\textbf{73.8}} &  \\
          & \multicolumn{1}{l}{SIGnet \cite{tnnls-2022}} & \multicolumn{1}{c}{96.2} & \multicolumn{1}{c}{86.8} & \multicolumn{1}{c}{{N/A}} &  \\
          & \multicolumn{1}{l}{Attention-Pred \cite{kbs-2022}} & \multicolumn{1}{c}{95.4} & \multicolumn{1}{c}{86.0} & \multicolumn{1}{c}{71.4} &  \\

\cmidrule{2-5}   & \multicolumn{1}{l}{LGN-Net(Ours) } & \multicolumn{1}{c}{\textbf{97.1}} & \multicolumn{1}{c}{\textbf{89.3}} & \multicolumn{1}{c}{73.0} &  \\
\cmidrule{1-5}          &       &       &       &       &  \\
    \end{tabular}%}
    }
  \label{tab:1}%
\end{table*}%

Prediction-based methods are usually more adept at exploiting spatiotemporal information among frames than reconstruction-based methods. Therefore, as shown in Table~\ref{tab:1}, the prediction-based methods are superior to reconstruction-based methods generally. Compared with reconstruction-based methods, LGN-Net achieved significantly superior performance as well.

The prediction-based methods have made lots of efforts to learn normality from training videos. However, all these methods have difficulty taking into account both local and global normality while favoring one of them. MNAD \cite{park2020learning} utilizes a memory module to memorize global prototype features and almost ignore local spatiotemporal representation. This causes MNAD \cite{park2020learning} to have difficulty representing diverse and complex normal patterns in ShanghaiTech \cite{luo2017revisit} dataset and achieves 70.5\% AUC only. Compared with MNAD  \cite{park2020learning}, LGN-Net capture local spatiotemporal representations using the spatiotemporal branch, which overcomes the shortcoming of MNAD. Therefore, it achieves remarkably superior performance (2.5\% AUC gain) compared with  MNAD \cite{park2020learning} on ShanghaiTech. Although AMMC-Net \cite{ammc} achieves a 0.7\% AUC gain on ShanghaiTech compared to LGN-Net, it suffers a 2.7\% AUC decrease on CUHK Avenue \cite{lu2013abnormal}.   In addition, AMMC-Net needs to calculate optical flow as temporal information using external networks.
\begin{figure*}[t]
  \centering
  \includegraphics[width=.95\textwidth]{ 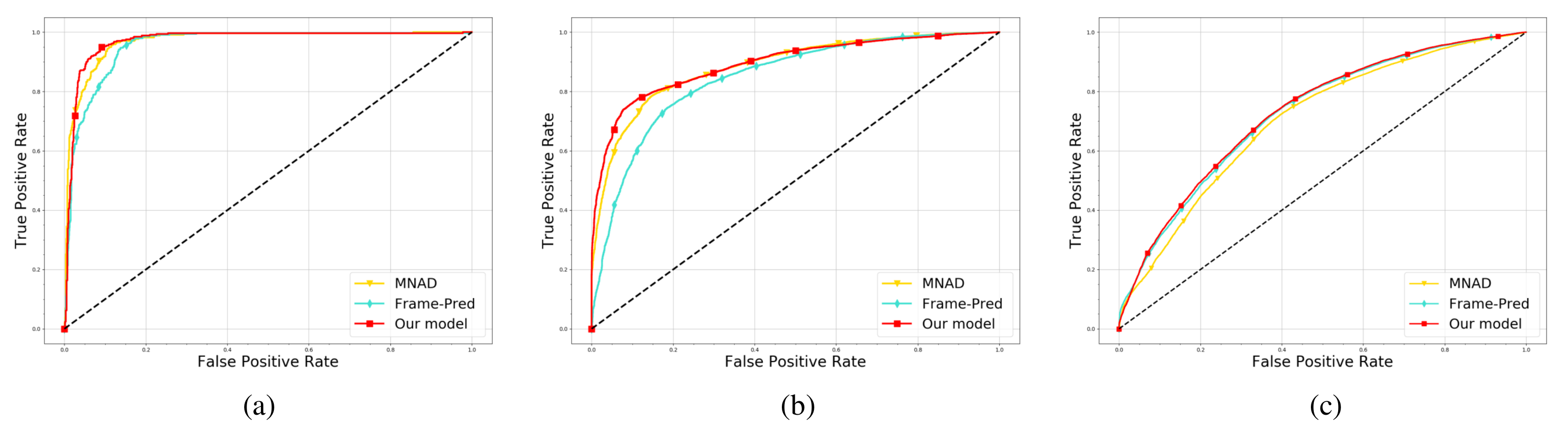}
  \caption{Frame-level ROC curves of our model, Frame-Pred \cite{liu2018future} and MNAD \cite{park2020learning} on UCSD Ped2 (a), CUHK Avenue (b), and ShanghaiTech (c) datasets.
}
  \label{fig:roc}
\end{figure*}

Frame-Pred \cite{liu2018future}, Multispace  \cite{tcsv2021}, STAN \cite{icassp} and STCEN \cite{pr-consistency}  over-focus on local normality. They tend to improve models' prediction ability by exploiting local spatial and temporal information and utilize adversarial training \cite{goodfellow2014generative} to enhance the learned local normality. The learned local normality enables them to represent more complex normal patterns. Thus Frame-Pred \cite{liu2018future}, Multispace  \cite{tcsv2021}, and  STCEN \cite{pr-consistency} achieve remarkable performance on ShanghaiTech. However, their strong representation ability allows them to represent some simple abnormal patterns well potentially \cite{gong2019memorizing}, leading to their relatively poor performance on UCSD ped2 and CUHK Avenue. Compared to these methods, LGN-Net can limit its representation capability for the simple abnormal instances utilizing recorded normal prototype features and therefore achieves remarkably superior performance on UCSD ped2 and CUHK Avenue. 

In Fig.~\ref{fig:roc}, we compared our model with other models in terms of the frame-level ROC curves. The comparison demonstrates that our model achieves superior performance compared to Frame-Pred\cite{liu2018future} on UCSD ped2 and CUHK Avenue. In addition, the comparison also indicates that our model achieves significantly superior performance compared to MNAD\cite{park2020learning} on ShanghaiTech. Since the codes of some recent methods are unavailable now, we choose these two methods for the comparison.
Frame-Pred and MNAD are well known and have excellent performance in the VAD task, and they are over-focusing on local and global normality, respectively.

In conclusion, compared with the methods in Table~\ref{tab:1}, LGN-Net is more adaptable to various scenes with different complexity. It shows the best performance on UCSD ped2 and CUHK Avenue while also achieving competitive performance in the more complex ShanghaiTech.

\subsection{Visualization Analysis}
In order to further demonstrate the effectiveness of our method on the VAD task, we provide the visualization results of LGN-Net. In addition, the visualization results could help our model to infer the time and position of abnormal events. 

\begin{figure*}[tp]
  \centering  \includegraphics[width=.92\textwidth]{ 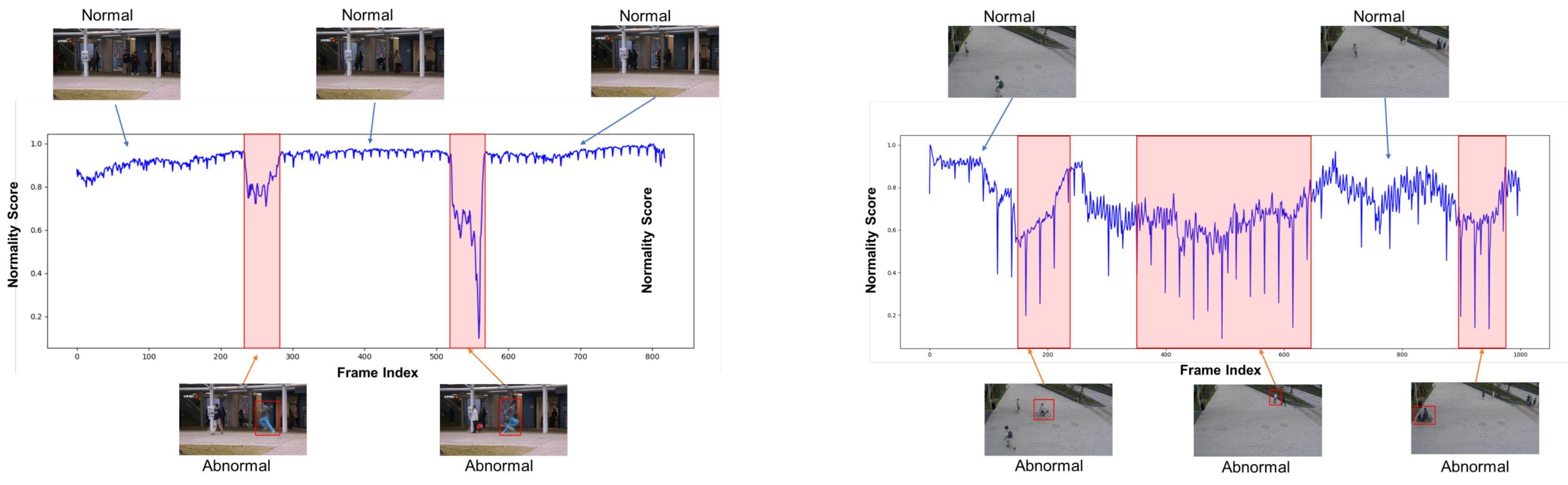}
  \caption{Normality score on two testing videos from CUHK Avenue (left)
and ShanghaiTech (right) datasets. The periods of abnormal events are marked by red regions. In CUHK Avenue, these two abnormal events are fast-running. In ShanghaiTech, the three abnormal events are bicycle riding, skateboarding, and motorcycle riding in the temporal order.
 }
  \label{fig:vis-all}
\end{figure*}
\begin{figure*}[bp]
  \centering
  \includegraphics[width=0.92\textwidth]{ 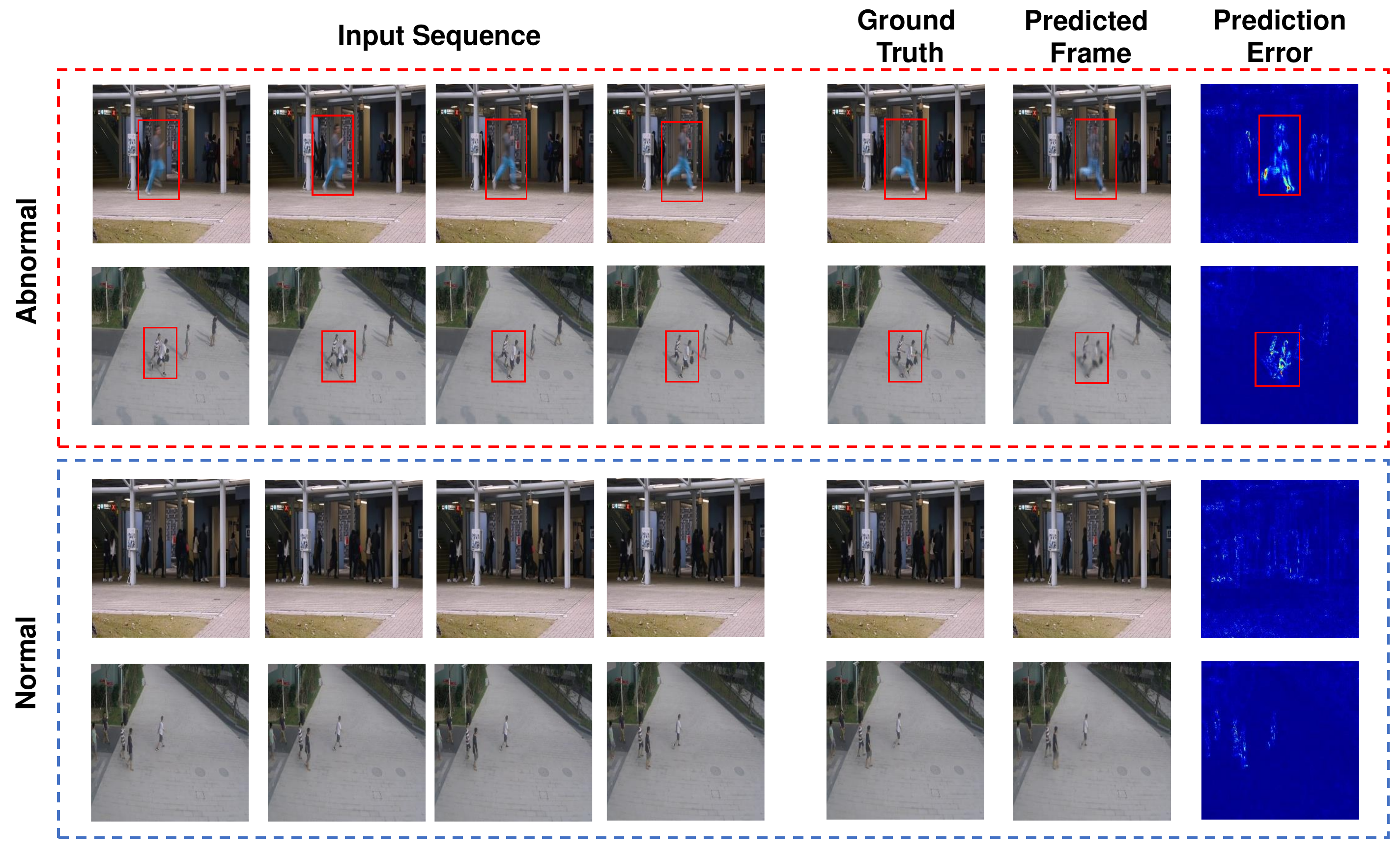}  
  \caption{Visualization of predicted frames and prediction errors. The first and third rows are the visualization results on CUHK Avenue, while the second and fourth rows are on ShanghaiTech. The abnormal events in the first and second rows are fast running and fighting, respectively. The red bounding boxes are used to mark the positions of abnormal events. The brighter color in the rightmost column denotes the larger prediction error.}
  \label{fig:vis-loc}
\end{figure*}

In Fig.~\ref{fig:vis-all}, we visualize normality scores output by LGN-Net on two testing videos from CUHK Avenue and ShanghaiTech. As mentioned in Section III-E, the higher normality score means the lower prediction error, and the lower score means the higher error. As shown in Fig.~\ref{fig:vis-all}, the normality score remains stable and relatively higher during normal events, while it drops drastically in these two videos during abnormal events. Thus, these visualization results further demonstrate the effectiveness of LGN-Net for abnormal event detection and also help LGN-Net infer the periods in which abnormal events occur.

In Fig.~\ref{fig:vis-loc}, we visualize prediction errors output by LGN-Net on four instances from CUHK Avenue and ShanghaiTech. The prediction error is the difference between a predicted frame and its corresponding ground truth. As shown in Fig.~\ref{fig:vis-loc}, prediction errors are very obvious in abnormal events, corresponding to the upper part of the figure. However, the predicted frames are remarkably close to the ground truth in the normal events, and thus the prediction errors are quite slight, corresponding to the lower part of the figure. We use red bounding boxes to mark the most obvious parts of the prediction errors in the rightmost column, which are exactly the positions of abnormal events. Therefore, these visualization results further demonstrate the effectiveness of LGN-Net for abnormal event detection and help LGN-Net to infer the spatial positions of abnormal events.

% Table generated by Excel2LaTeX from sheet 'ablation'

\subsection{Ablation Studies}
We conduct the ablation analysis in this section to further demonstrate the effectiveness of considering both local and global normality. We report the frame-level AUCs of LGN-Net and its variants on CUHK Avenue and ShanghaiTech in Table~\ref{tab:2}. Loc-Net and Glo-Net are baseline models. Loc-Net includes only the spatiotemporal branch based on ConvLSTM \cite{xingjian2015convolutional}, which results in over-focusing on local normality. Glo-Net includes only the prototype branch similar to MNAD \cite{park2020learning}, which results in over-focusing on global normality. Similar to MNAD \cite{park2020learning}, we concatenate latent space features $\bm F_{lat}$ and the global features $\bm{F}_{glo}$ and feed them to the decoder $D_{lg}$ in Glo-Net.  LGN-ST contains the spatiotemporal branch and the prototype branch, but its STP-Net consist of ConvLSTM \cite{xingjian2015convolutional} rather than ST-LSTM \cite{wang2017predrnn}. In LGN-ST, we concatenate $\bm{H}_{t}^{4}$ and $\bm{C}_{t}^{4}$ as the final local spatiotemporal representations.   LGN-Net is our final proposed model, which contains the two branches and utilizes ST-LSTM  \cite{wang2017predrnn} as units of STP-Net.  

\begin{table*}[tp]
\footnotesize
  \centering
  \caption{Quantitative comparison for variants of our model. We measure the frame-level AUC (\%) on CUHK avenue and ShanghaiTech. Bold number indicates the best performance.}
    \setlength{\tabcolsep}{1.1mm}{
    \begin{tabular}{cccc|c|c}

    \hline
    Model & Spatiotemporal & Prototype  & ST-LSTM & CUHK Avenue  & ShanghaiTech \\
    \hline
    Loc-Net     & \Checkmark    & \XSolidBrush    & \XSolidBrush     & 84.7  & 71.4 \\
    Glo-Net     & \XSolidBrush    & \Checkmark     & \XSolidBrush    & 88.5  & 70.3 \\
    LGN-ST     & \Checkmark      & \Checkmark     & \XSolidBrush    & 88.8  & 72.1 \\
    \hline
LGN-Net(Ours) & \Checkmark    & \Checkmark    & \Checkmark    &\textbf{ 89.3 }   &\textbf{ 73.0} \\
    \hline
    \end{tabular}%
    }
  \label{tab:2}%
\end{table*}%

Compared with Loc-Net, Glo-Net achieves a 3.8\% AUC gain on CUHK Avenue while getting a 1.1\% AUC decrease on ShanghaiTech, further validating the global normality imposes a limitation on the representation for complex normal patterns. LGN-ST validates the effectiveness of local normality in diverse and complex patterns. It achieves a 1.8\% AUC gain on ShanghaiTech compared to Glo-Net. In addition, LGN-ST achieves a 4.1\% AUC gain on CUHK Avenue compared to Loc-Net. This also indicates the effectiveness of global normality for the simpler dataset.
LGN-Net achieves 0.5\% and 0.9\% AUC gains on CUHK Avenue and ShanghaiTech compared to LGN-ST, respectively,  which further shows the importance of local spatiotemporal representations for the VAD task. 
 \begin{figure}[ht]
  \centering
  \includegraphics[width=.65\textwidth]{ 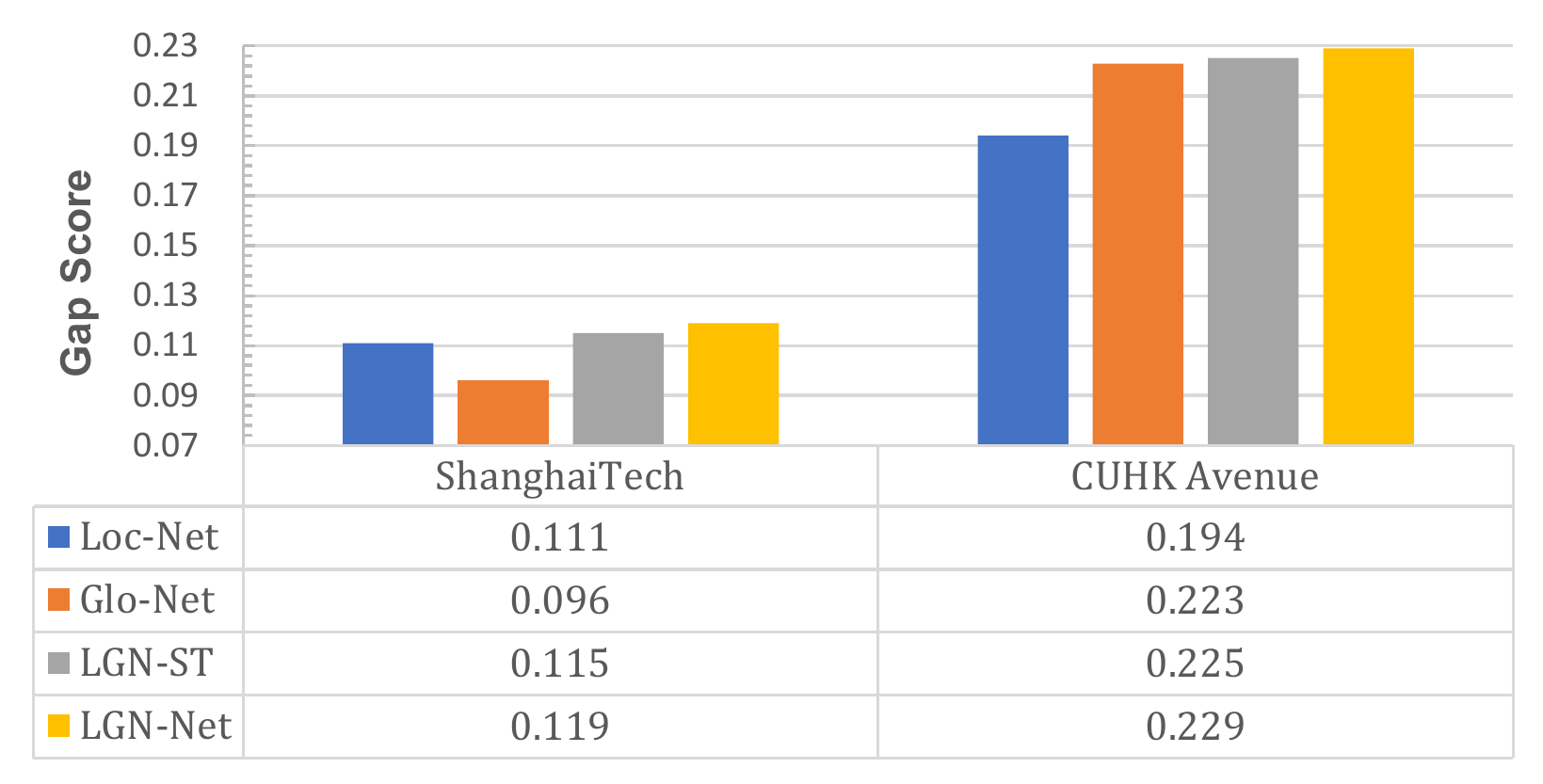}
  \caption{Gap score of LGN-Net and its variants on ShanghaiTech and CUHK Avenue. Gap score is the average normality score of normal frames minus that of the abnormal frames. The larger gap score indicates the better separability for normal and abnormal frames.}
  \label{fig:mean}
\end{figure}

 \begin{figure}[tp]
  \centering
  \includegraphics[width=.7\textwidth]{ 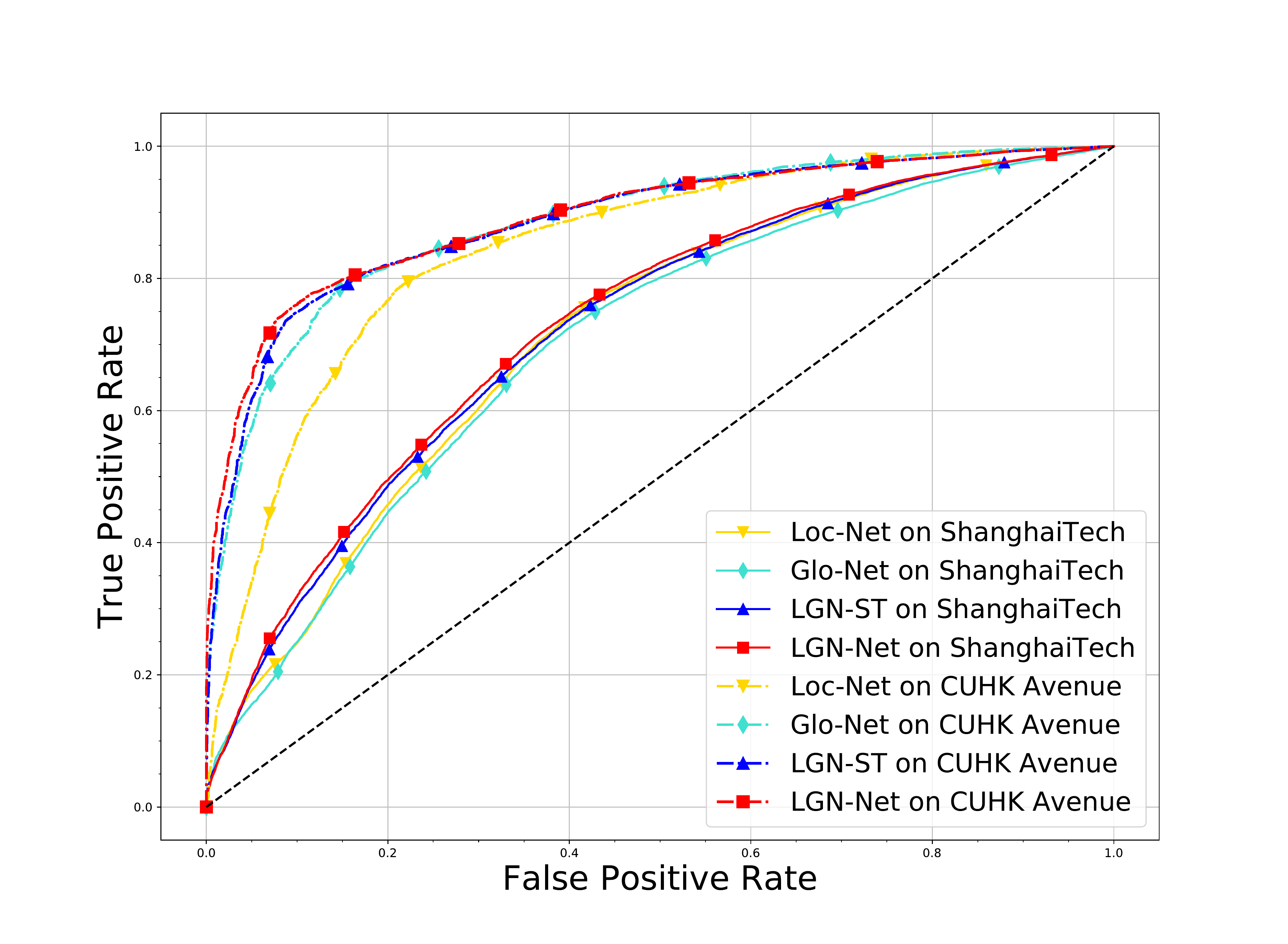}
  \caption{Frame-level ROC curves of LGN-Net and its variants on CUHK Avenue and ShanghaiTech dataset.}
  \label{fig:ablation_roc}
\end{figure}

 \begin{figure}[bp]
  \centering
  \includegraphics[width=.7\textwidth]{ 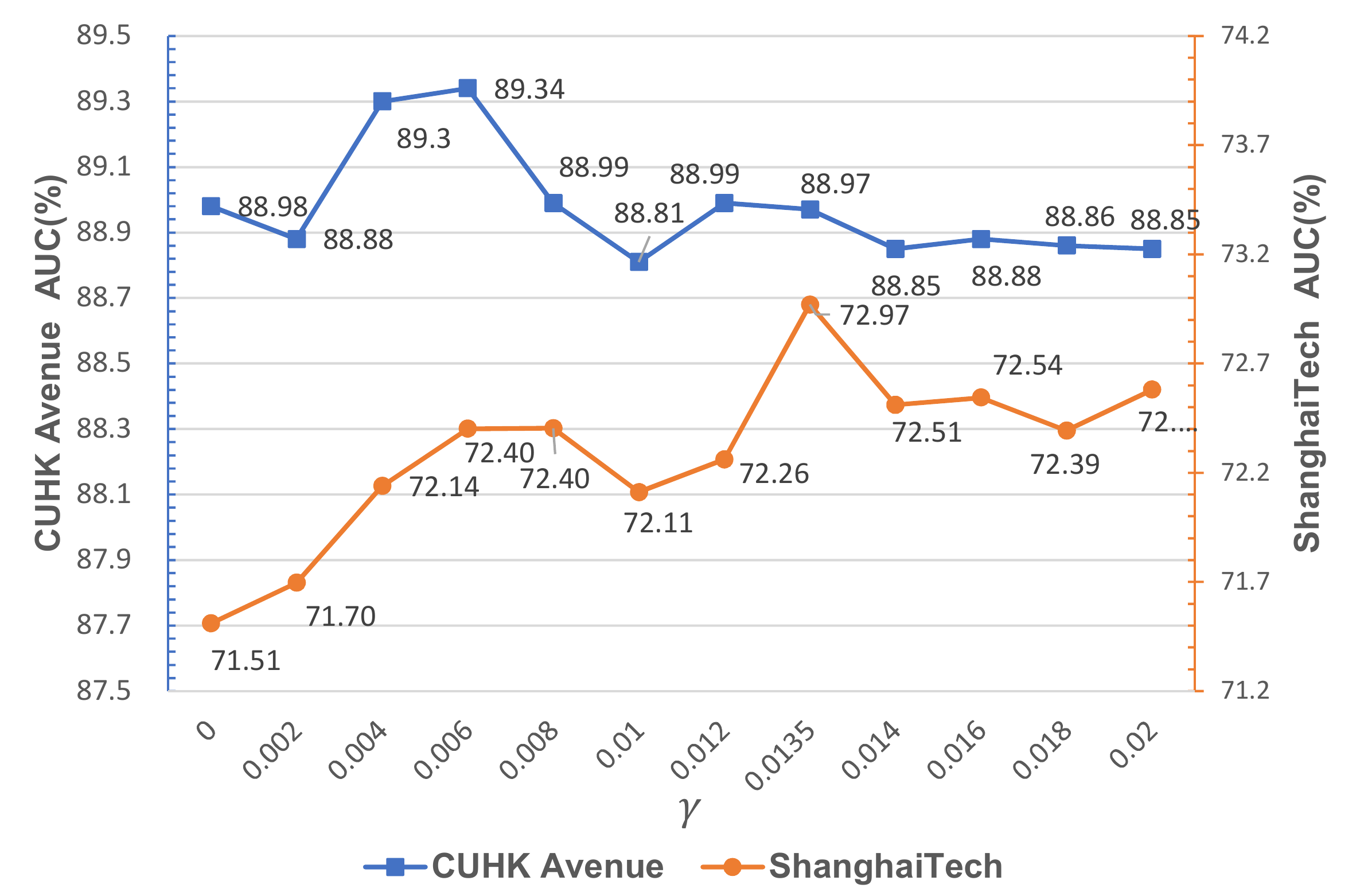}
  \caption{Frame-level AUC (\%) under different $\gamma$ on CUHK Avenue and ShanghaiTech datasets.}
  \label{fig:gamma}
\end{figure}
To further validate the effectiveness of considering both local and global normality, we compared the gap score and frame-level ROC curves of LGN-Net with its variants on CUHK Avenue and ShanghaiTech datasets. As shown in Fig.~\ref{fig:mean}, LGN-Net achieves 0.004 and 0.023 gains in terms of the gap score compared to LGN-ST and Glo-Net, respectively, which indicates that the local normality enables LGN-Net to have superior performance in separating normal and abnormal frames of ShanghaiTech dataset. Furthermore, compared to Loc-Net, LGN-Net achieves a 0.035 gain in terms of the gap score, indicating that the global normality helps LGN-Net separate normal and abnormal frames of CUHK Avenue.
In addition, Fig.~\ref{fig:ablation_roc} indicates the effectiveness of both local and global normality for the performance of LGN-Net on CUHK Avenue and ShanghaiTech datasets.

\subsection{Hyperparameter Analysis}

 \begin{figure}[htb]
  \centering
  \includegraphics[width=.7\textwidth]{ 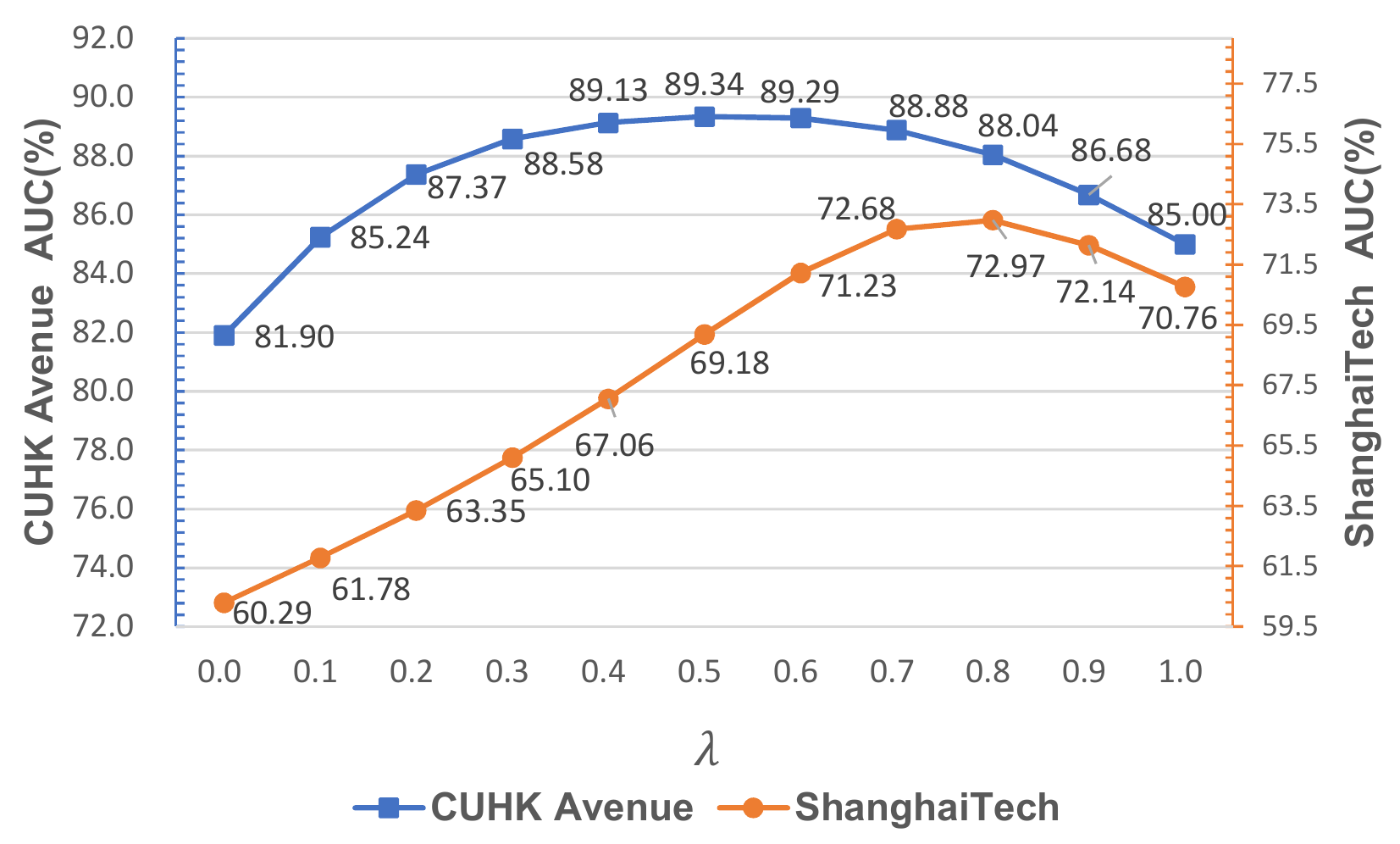}
  \caption{Frame-level AUC (\%) under different $\lambda$ on CUHK Avenue and ShanghaiTech  datasets.}
  \label{fig:lamda}
\end{figure}
We utilize a threshold $\gamma$ to select some normal frames from testing sets for updating the memory module in Section III-C and set a parameter $\lambda$ to balance PSNR and distance $D$ in Eq.~\ref{eqd}. In this Section, as shown in Fig.~\ref{fig:gamma} and Fig.~\ref{fig:lamda}, we report the AUC performance of LGN-Net with different $\gamma$ and $\lambda$ respectively on CUHK Avenue and ShanghaiTech. When we set $\gamma$ to 0, which means LGN-Net does not use frames in testing videos to update the memory module. This lead to the model achieving 0.36\% and 1.46\% AUC decreases on CUHK Avenue and ShanghaiTech compared with the best performances, respectively. When we set $\lambda$ to 1, which means we only utilize PSNR to calculate the normality score. In this situation, the AUCs decreased by 4.34\% and 1.92\% on CUHK Avenue and ShanghaiTech compared with the best performances, respectively.
Therefore, these comparisons show the influence of parameters $\gamma$ and $\lambda$ on our model's performance and validate the effectiveness of these two strategies.

\section{Conclusion}
In this paper, we focus on the issue that most existing VAD methods over-focus only on local or global normality, making them difficult to balance the representation for normal and abnormal patterns and generalize to different scenes. To address this issue, we introduce a novel unsupervised VAD method to consider both local and global normality. We propose a two-branch model named LGN-Net, where one branch learns spatiotemporal evolution regularities in consecutive frames as local normality, while the other branch memorizes prototypical patterns of all normal videos as global normality. The local normality enables LGN-Net to understand more diverse and complex normal patterns, while the global normality can limit LGN-Net's representation for abnormal instances. We fuse the local and global normality to enable LGN-Net more adaptable to various scenes with different complexity. The experimental results show the superiority of our method while demonstrating the effectiveness of considering both local and global normality.  In future work, we will explore a better way to consider local and global normality in the VAD task.
\section{Acknowledgment}
This work was supported in part by the Science and Technology Commission of Shanghai Municipality Research Fund (Grant No. 21JC1405300), in part by the National Key Research and Development Program of China (Grant No. 2018YFC0831102), and in part by the Shanghai Key Research Lab. of NSAI.



\bibliographystyle{elsarticle-num} 
\bibliography{sample}

\begin{thebibliography}{10}
\expandafter\ifx\csname url\endcsname\relax
  \def\url#1{\texttt{#1}}\fi
\expandafter\ifx\csname urlprefix\endcsname\relax\def\urlprefix{URL }\fi
\expandafter\ifx\csname href\endcsname\relax
  \def\href#1#2{#2} \def\path#1{#1}\fi

\bibitem{kbs-echo}
W.~Ullah, T.~Hussain, Z.~A. Khan, U.~Haroon, S.~W. Baik, Intelligent dual
  stream cnn and echo state network for anomaly detection, Knowl.-Based Syst.
  253 (2022) 109456.
\newblock \href {https://doi.org/https://doi.org/10.1016/j.knosys.2022.109456}
  {\path{doi:https://doi.org/10.1016/j.knosys.2022.109456}}.

\bibitem{kbs-2022}
Q.~Li, R.~Yang, F.~Xiao, B.~Bhanu, F.~Zhang, Attention-based anomaly detection
  in multi-view surveillance videos, Knowl.-Based Syst. 252 (2022) 109348.
\newblock \href {https://doi.org/https://doi.org/10.1016/j.knosys.2022.109348}
  {\path{doi:https://doi.org/10.1016/j.knosys.2022.109348}}.

\bibitem{muti-task}
M.-I. Georgescu, A.~Bărbălău, R.~T. Ionescu, F.~Shahbaz~Khan, M.~Popescu,
  M.~Shah, Anomaly detection in video via self-supervised and multi-task
  learning, in: Proc. IEEE/CVF Conf. Comput. Vis. Pattern Recognit., 2021, pp.
  12737--12747.

\bibitem{pami2021}
M.~I. Georgescu, R.~T. Ionescu, F.~S. Khan, M.~Popescu, M.~Shah, A
  background-agnostic framework with adversarial training for abnormal event
  detection in video, {IEEE} Trans. Pattern Anal. Mach. Intell. 44~(9) (2022)
  4505--4523.
\newblock \href {https://doi.org/10.1109/TPAMI.2021.3074805}
  {\path{doi:10.1109/TPAMI.2021.3074805}}.

\bibitem{insai}
L.~Song, X.~Hu, G.~Zhang, P.~Spachos, K.~N. Plataniotis, H.~Wu, Networking
  systems of ai: On the convergence of computing and communications, IEEE
  Internet Things J. 9~(20) (2022) 20352--20381.

\bibitem{liu-pami}
W.~Luo, W.~Liu, D.~Lian, J.~Tang, L.~Duan, X.~Peng, S.~Gao, Video anomaly
  detection with sparse coding inspired deep neural networks, {IEEE} Trans.
  Pattern Anal. Mach. Intell. 43~(3) (2021) 1070--1084.

\bibitem{Anopcn}
M.~Ye, X.~Peng, W.~Gan, W.~Wu, Y.~Qiao, Anopcn: Video anomaly detection via
  deep predictive coding network, in: Proc. 27th ACM Int. Conf. Multimedia,
  2019, p. 1805–1813.

\bibitem{chen2021nm}
D.~Chen, L.~Yue, X.~Chang, M.~Xu, T.~Jia, Nm-gan: Noise-modulated generative
  adversarial network for video anomaly detection, Pattern Recognit. 116 (2021)
  107969.

\bibitem{tmm-recon}
Z.~Fang, J.~T. Zhou, Y.~Xiao, Y.~Li, F.~Yang, Multi-encoder towards effective
  anomaly detection in videos, {IEEE} Trans. Multimedia 23 (2021) 4106--4116.

\bibitem{liu2022collaborative}
Y.~Liu, J.~Liu, M.~Zhao, S.~Li, L.~Song, Collaborative normality learning
  framework for weakly supervised video anomaly detection, IEEE Trans. Circuits
  and Syst. II, Exp. Briefs 69~(5) (2022) 2508--2512.
\newblock \href {https://doi.org/10.1109/TCSII.2022.3161061}
  {\path{doi:10.1109/TCSII.2022.3161061}}.

\bibitem{liu2022learning}
Y.~Liu, J.~Liu, X.~Zhu, D.~Wei, X.~Huang, L.~Song, Learning task-specific
  representation for video anomaly detection with spatial-temporal attention,
  in: Proc. IEEE Int. Conf. Acoust. Speech Signal Process., 2022, pp.
  2190--2194.

\bibitem{liu2022appearance}
Y.~Liu, J.~Liu, J.~Lin, M.~Zhao, L.~Song, Appearance-motion united auto-encoder
  framework for video anomaly detection, IEEE Trans. Circuits and Syst. II,
  Exp. Briefs 69~(5) (2022) 2498--2502.

\bibitem{liu2018future}
W.~Liu, W.~Luo, D.~Lian, S.~Gao, Future frame prediction for anomaly
  detection--a new baseline, in: Proc. IEEE/CVF Conf. Comput. Vis. Pattern
  Recognit., 2018, pp. 6536--6545.

\bibitem{icassp}
S.~Lee, H.~G. Kim, Y.~M. Ro, Stan: Spatio-temporal adversarial networks for
  abnormal event detection, in: Proc. IEEE Int. Conf. Acoust. Speech Signal
  Process., 2018, pp. 1323--1327.

\bibitem{pr-consistency}
Y.~Hao, J.~Li, N.~Wang, X.~Wang, X.~Gao, Spatiotemporal consistency-enhanced
  network for video anomaly detection, Pattern Recognit. 121 (2022) 108232.

\bibitem{tcsv2021}
Y.~Zhang, X.~Nie, R.~He, M.~Chen, Y.~Yin, Normality learning in multispace for
  video anomaly detection, {IEEE} Trans. Circuits Syst. Video Technol. 31~(9)
  (2021) 3694--3706.

\bibitem{unet}
O.~Ronneberger, P.~Fischer, T.~Brox, U-net: Convolutional networks for
  biomedical image segmentation, in: Proc. Int. Conf. Med. Image Comput.
  Comput.-Assisted Intervention, Springer, 2015, pp. 234--241.

\bibitem{park2020learning}
H.~Park, J.~Noh, B.~Ham, Learning memory-guided normality for anomaly
  detection, in: Proc. IEEE/CVF Conf. Comput. Vis. Pattern Recognit., 2020, pp.
  14372--14381.

\bibitem{ammc}
R.~Cai, H.~Zhang, W.~Liu, S.~Gao, Z.~Hao, Appearance-motion memory consistency
  network for video anomaly detection, in: Proc. AAAI Conf. Artif. Intell.,
  2021, pp. 938--946.

\bibitem{wang2017predrnn}
Y.~Wang, M.~Long, J.~Wang, Z.~Gao, P.~S. Yu, Predrnn: Recurrent neural networks
  for predictive learning using spatiotemporal lstms, in: Proc. 31th Int. Conf.
  Neural Inf. Process. Syst., 2017, pp. 879--888.

\bibitem{wang2021predrnn}
Y.~Wang, H.~Wu, J.~Zhang, Z.~Gao, J.~Wang, P.~Yu, M.~Long, Predrnn: A recurrent
  neural network for spatiotemporal predictive learning, {IEEE} Trans. Pattern
  Anal. Mach. Intell. (2022) 1--1.

\bibitem{hybrid}
Z.~Liu, Y.~Nie, C.~Long, Q.~Zhang, G.~Li, A hybrid video anomaly detection
  framework via memory-augmented flow reconstruction and flow-guided frame
  prediction, in: Proc. IEEE/CVF Int. Conf. Comput. Vis., 2021, pp.
  13568--13577.

\bibitem{tcsvt2020}
J.~Guo, P.~Zheng, J.~Huang, Efficient privacy-preserving anomaly detection and
  localization in bitstream video, {IEEE} Trans. Circuits Syst. Video Technol.
  30~(9) (2020) 3268--3281.

\bibitem{gong2019memorizing}
D.~Gong, L.~Liu, V.~Le, B.~Saha, M.~R. Mansour, S.~Venkatesh, A.~v.~d. Hengel,
  Memorizing normality to detect anomaly: Memory-augmented deep autoencoder for
  unsupervised anomaly detection, in: Proc. IEEE/CVF Int. Conf. Comput. Vis.,
  2019, pp. 1705--1714.

\bibitem{lmc-net}
S.~Lee, H.~G. Kim, D.~H. Choi, H.~Kim, Y.~M. Ro, Video prediction recalling
  long-term motion context via memory alignment learning, in: Proc. IEEE/CVF
  Conf. Comput. Vis. Pattern Recognit., 2021, pp. 3054--3063.

\bibitem{DBLP:conf/icmcs/LaiLH20}
Y.~Lai, R.~Liu, Y.~Han, Video anomaly detection via predictive autoencoder with
  gradient-based attention, in: Proc. IEEE Int. Conf. Multimedia Expo, 2020,
  pp. 1--6.

\bibitem{hasan2016learning}
M.~Hasan, J.~Choi, J.~Neumann, A.~K. Roy-Chowdhury, L.~S. Davis, Learning
  temporal regularity in video sequences, in: Proc. IEEE/CVF Conf. Comput. Vis.
  Pattern Recognit., 2016, pp. 733--742.

\bibitem{luo2017revisit}
W.~Luo, W.~Liu, S.~Gao, A revisit of sparse coding based anomaly detection in
  stacked rnn framework, in: Proc. IEEE Int. Conf. Comput. Vis., 2017, pp.
  341--349.

\bibitem{abnormalgan}
M.~Ravanbakhsh, M.~Nabi, E.~Sangineto, L.~Marcenaro, C.~Regazzoni, N.~Sebe,
  Abnormal event detection in videos using generative adversarial nets, in:
  Proc. IEEE Int. Conf. Image Process., 2017, pp. 1577--1581.

\bibitem{goodfellow2014generative}
I.~J. Goodfellow, J.~Pouget-Abadie, M.~Mirza, B.~Xu, D.~Warde-Farley, S.~Ozair,
  A.~C. Courville, Y.~Bengio, Generative adversarial nets, in: Proc. 27th Int.
  Conf. Neural Inf. Process. Syst., 2014, pp. 2672–--2680.

\bibitem{memory-network}
J.~Weston, S.~Chopra, A.~Bordes, Memory networks, in: Proc. Int. Conf. Learn.
  Representations, 2015.

\bibitem{xingjian2015convolutional}
S.~Xingjian, Z.~Chen, H.~Wang, D.-Y. Yeung, W.-K. Wong, W.-c. Woo,
  Convolutional lstm network: A machine learning approach for precipitation
  nowcasting, in: Proc. 28th Int. Conf. Neural Inf. Process. Syst., 2015, pp.
  802--810.

\bibitem{chang2022video}
Y.~Chang, Z.~Tu, W.~Xie, B.~Luo, S.~Zhang, H.~Sui, J.~Yuan, Video anomaly
  detection with spatio-temporal dissociation, Pattern Recognit. 122 (2022)
  108213.

\bibitem{ramachandra2020survey}
B.~Ramachandra, M.~J. Jones, R.~R. Vatsavai, A survey of single-scene video
  anomaly detection, {IEEE} Trans. Pattern Anal. Mach. Intell. 44~(5) (2022)
  2293--2312.

\bibitem{nguyen_anomaly_2020}
K.-T. Nguyen, D.-T. Dinh, M.~N. Do, M.-T. Tran, Anomaly {Detection} in
  {Traffic} {Surveillance} {Videos} with {GAN}-based {Future} {Frame}
  {Prediction}, in: Proc. Int. Conf. Multimedia Retrieval, Association for
  Computing Machinery, 2020, pp. 457--463.

\bibitem{PSNR}
M.~Mathieu, C.~Couprie, Y.~Lecun, Deep multi-scale video prediction beyond mean
  square error, in: Proc. Int. Conf. Learn. Representations, 2016.

\bibitem{mahadevan2010anomaly}
V.~Mahadevan, W.~Li, V.~Bhalodia, N.~Vasconcelos, Anomaly detection in crowded
  scenes, in: Proc. IEEE Conf. Comput. Vis. Pattern Recognit., 2010, pp.
  1975--1981.

\bibitem{lu2013abnormal}
C.~Lu, J.~Shi, J.~Jia, Abnormal event detection at 150 fps in matlab, in: Proc.
  IEEE Int. Conf. Comput. Vis., 2013, pp. 2720--2727.

\bibitem{tmm-crowded}
N.~Li, F.~Chang, C.~Liu, Spatial-temporal cascade autoencoder for video anomaly
  detection in crowded scenes, {IEEE} Trans. Multimedia 23 (2021) 203--215.
\newblock \href {https://doi.org/10.1109/TMM.2020.2984093}
  {\path{doi:10.1109/TMM.2020.2984093}}.

\bibitem{tmm-end}
K.~Xu, T.~Sun, X.~Jiang, Video anomaly detection and localization based on an
  adaptive intra-frame classification network, {IEEE} Trans. Multimedia 22~(2)
  (2020) 394--406.

\bibitem{kingma2014adam}
D.~P. Kingma, J.~Ba, Adam: A method for stochastic optimization, in: Proc. Int.
  Conf. Learn. Representations, 2015.

\bibitem{kim2009observe}
J.~Kim, K.~Grauman, Observe locally, infer globally: a space-time mrf for
  detecting abnormal activities with incremental updates, in: Proc. IEEE Conf.
  Comput. Vis. Pattern Recognit., 2009, pp. 2921--2928.

\bibitem{luo2017remembering}
W.~Luo, W.~Liu, S.~Gao, Remembering history with convolutional lstm for anomaly
  detection, in: Proc. IEEE Int. Conf. Multimedia Expo, 2017, pp. 439--444.

\bibitem{nguyen2019anomaly}
T.-N. Nguyen, J.~Meunier, Anomaly detection in video sequence with
  appearance-motion correspondence, in: Proc. IEEE/CVF Int. Conf. Comput. Vis.,
  2019, pp. 1273--1283.

\bibitem{tcsvt2019}
J.~T. Zhou, L.~Zhang, Z.~Fang, J.~Du, X.~Peng, Y.~Xiao, Attention-driven loss
  for anomaly detection in video surveillance, {IEEE} Trans. Circuits Syst.
  Video Technol. 30~(12) (2020) 4639--4647.

\bibitem{stc-net}
M.~Zhao, Y.~Liu, J.~Liu, X.~Zeng, Exploiting spatial-temporal correlations for
  video anomaly detection, in: Proc. IEEE/CVF Int. Conf. Pattern Recognit.,
  2022, pp. 1727--1733.
\newblock \href {https://doi.org/10.1109/ICPR56361.2022.9956287}
  {\path{doi:10.1109/ICPR56361.2022.9956287}}.

\bibitem{tnnls-2022}
Z.~Fang, J.~Liang, J.~T. Zhou, Y.~Xiao, F.~Yang, Anomaly detection with
  bidirectional consistency in videos, IEEE Trans. Neural Netw. Learn. Syst.
  33~(3) (2022) 1079--1092.
\newblock \href {https://doi.org/10.1109/TNNLS.2020.3039899}
  {\path{doi:10.1109/TNNLS.2020.3039899}}.

\end{thebibliography}




%%\bibitem{}

%%\end{thebibliography}
\end{document}